# Physics of Language Models: Part 3.2, Knowledge Manipulation

Zeyuan Allen-Zhu
zeyuanallenzhu@meta.com
Meta AI / FAIR Labs

Yuanzhi Li
Yuanzhi.Li@mbzuai.ac.ae
Mohamed bin Zayed University of AI

September 18, 2023

(version 2)*

**Abstract**

Language models can store vast factual knowledge, yet their ability to flexibly use this knowledge for downstream tasks (e.g., via instruction finetuning) remains questionable. This paper investigates four fundamental knowledge manipulation tasks: **retrieval** (e.g., "What is person A's attribute X?"), **classification** (e.g., "Is A's attribute X even or odd?"), **comparison** (e.g., "Is A greater than B in attribute X?"), and **inverse search** (e.g., "Which person's attribute X equals T?").

We show that language models excel in knowledge retrieval but struggle even in the simplest classification or comparison tasks unless Chain of Thoughts (CoTs) are employed during both training and inference. Moreover, their performance in inverse knowledge search is virtually 0%, regardless of the prompts. Our primary contribution is a *controlled, synthetic experiment* that confirms these weaknesses are *inherent* to language models: they cannot efficiently manipulate knowledge from pre-training data, even when such knowledge is perfectly stored in the models, despite adequate training and sufficient model size. Our findings also apply to modern pretrained language models such as GPT-4, thus giving rise to many Turing tests to distinguish Humans from contemporary AIs.

---

*Project page: https://physics.allen-zhu.com/part-3-knowledge/part-3-2. An extended video of this paper is available at https://youtu.be/YSHzKmEianc. V1 was circulated internally at Meta on Sep 18, 2023, and appeared on arXiv on Sep 25, 2023. V2 polishes writing and includes additional Llama/Mistral experiments and larger data; but the conclusions remain unchanged.

We would like to thank Lin Xiao, Chunting Zhou, Xiaodong Liu, Zhijie Zhou for many helpful conversations. We would like to extend special thanks to Nabib Ahmed, Giri Anantharaman, Lucca Bertoncini, Henry Estela, Liao Hu, Caleb Ho, Wil Johnson, Apostolos Kokolis, and Shubho Sengupta from Meta FAIR, as well as Ian Clark, Gourab De, Anmol Mann, and Max Pfeifer from W&B; without their invaluable support, the experiments in this paper would not have been possible.

# 1 Introduction

Knowledge is a fundamental component of human civilization and intelligence. Throughout our lives, we accumulate a vast amount of knowledge and learn to use it flexibly. Large language models like GPT-4 [23] have demonstrated an impressive capacity to memorize knowledge, arguably surpassing any human. These models also show signs of being able to manipulate this knowledge to solve various problems.

In this work, we aim to understand how transformer-based language models manipulate the knowledge they have memorized during pretraining and use it flexibly to solve different tasks at inference time. For example, can language models determine if Princeton is ranked higher than MIT based on its stored 2023 US News university ranking knowledge? Can they answer questions such as "Was Joe Biden born in an odd year?" or "Was Donald Trump born earlier than Nancy Pelosi?" based on their memorization of celebrities' birthdays? (Spoiler alert, even GPT-4 or Llama-3 *still* fail to answer these as of May 10, 2024, see Figure 9; this paper explains why.)

In other words, we are interested in questions that are *functions* of specific knowledge from the pretraining data, and study a language model's ability to answer questions during inference time. Knowledge manipulation is arguably *a simplest form of logical reasoning*. To answer questions like "Is Person A's attribute X good?", a model not previously exposed to this sentence in its training data may draw conclusions from other data such as "Person A's attribute X equals T" and "T is good".

In this paper, "knowledge" refers to *factual knowledge* (e.g., knowledge graph), and we explore whether a language model can logically manipulate such knowledge embedded in its model weights. Other research may focus on in-context knowledge or RAG [6, 14, 15, 17–19, 24, 29, 32], where the model responds to queries about a *provided paragraph* in the context (possibly via RAG).

Extensive research has been conducted on the question-answering capabilities of language models at inference time [11, 20, 22, 25, 26, 30, 31, 34], primarily focusing on models trained with internet data. A significant challenge in determining whether these models can manipulate knowledge is to ascertain if the internet data already contains the exact or equivalent question, or if the models genuinely performed logical deduction during inference time.

We are particularly interested in scenarios *without data contamination*: the questions or their equivalent forms should not appear in the model's training data, while the same "function" for other knowledge should be present — thus ensuring the model understands the function. For example, can the model determine "Was Joe Biden born in an odd year?" if it has not encountered this sentence or its equivalents during pretraining (such as "Is Joe Biden's birth year divisible by 2"), but can infer from "Biden was born in 1942" and "1942 is not odd"? Answering such questions requires the model to both memorize and comprehend the knowledge. (See Figure 1.)

To address the *unpredictability of internet data*, Allen-Zhu and Li [2, 3] developed synthetic pretrain data containing controlled biographies for up to $N = 20$ million individuals. They explored how a language model stores and extracts knowledge about these individuals after-pretraining. Here is an example of their biography data:

> Anya Briar Forger was born on October 2, 1996. She spent her early years in Princeton, NJ. She received mentorship and guidance from faculty members at Massachusetts Institute of Technology. She completed her education with a focus on Communications. She had a professional role at Meta Platforms. She was employed in Menlo Park, CA.

(1.1)

Allen-Zhu and Li [2] found that a pretrained model may struggle to *extract* stored knowledge from biographical data unless the data is sufficiently *knowledge-augmented*, meaning the same biography has diverse and well-permuted English descriptions (see Section 2). This augmentation aids in accurately answering extraction queries such as "Which city was Anya Briar Forger born in?"

While we recommend reading our concurrent work [2] first, this paper can be read independently.

## 1.1 Our Results

This paper further explores whether a model, pre-trained on augmented biography data, can *manipulate* its knowledge after instruction finetuning. We investigate its ability to handle queries that require reasoning about personal attributes, such as "Was Anya born in a southern city?" or "Is Anya's university better than Sabrina's?"

During training, the model learns from the biographies of all $N$ individuals and the knowledge manipulation question-answer (QA) texts from a subset of individuals (the in-distribution set $\mathcal{P}_{\mathsf{train}}$). We evaluate the model's *out-of-distribution* (OOD) generation accuracy by testing it on the remaining subset (the out-of-distribution set $\mathcal{P}_{\mathsf{test}}$), where it has seen the biographies but not the QAs during training. Including $\mathcal{P}_{\mathsf{train}}$ in the training data ensures the model encounters enough examples to comprehend the QAs. We focus on the model's OOD accuracy on $\mathcal{P}_{\mathsf{test}}$, reflecting its true capability in logical deduction during inference time, as opposed to on $\mathcal{P}_{\mathsf{train}}$ which could easily reach 100%.

We study four basic types of knowledge manipulations: retrieval, classification, comparison, and inverse search, which cover most real-world scenarios.[1]

KNOWLEDGE RETRIEVAL. Extending work on knowledge extraction [2], we finetune the model to retrieve (1) part of an attribute or (2) multiple attributes at once. We discover a model may

- correctly answer "What is the birth date of Anya" as "June 27th, 1997", but struggle with "What is the birth year of Anya" (**Result 2**); and
- correctly answer "Which company and where did Anya work" but fail on "Where and which company did Anya work." (**Result 1**)

These serve as **preliminary evidence** suggesting the necessity of a Chain-of-Thought (CoT) for knowledge manipulation. The model must *explicitly state* the birth month/day to deduce the birth year, or *explicitly state* the company name before the work city location.

KNOWLEDGE CLASSIFICATION. We finetune the model for classification tasks on its stored knowledge; for instance, "What degree did Anya receive?" may require ternary classification (art, science, engineering) based on her major. Language models often struggle with such tasks unless they (1) generate answers in CoT manner or (2) are finetuned with a significantly larger number of samples than theoretically necessary.

---

[1]One could also explore combinations, such as "Is A's wife's university ranked higher than B's?" or "Is the person born on June 27th, 1997, and studied at MIT named with an initial A?" These would further complicate the tasks. Given that we show mostly negative results, focusing on the basic forms suffices.

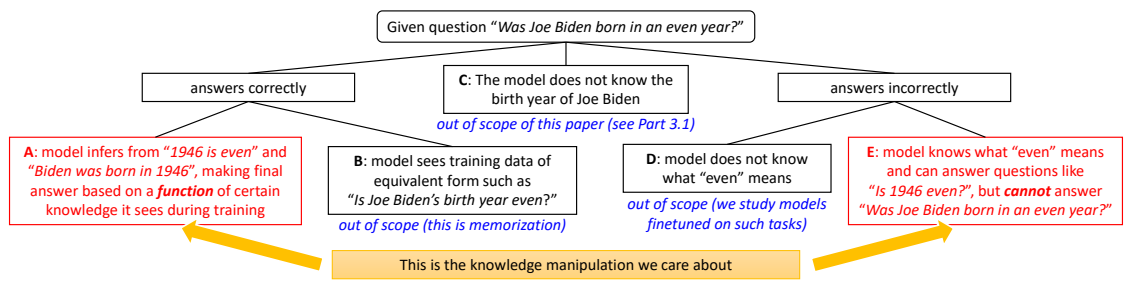

Figure 1: We study (A) vs (E) as knowledge manipulation. With a pre-trained model over internet data, it is very hard to determine whether (B,C,D) has happened due to the uncontrollability of internet data.

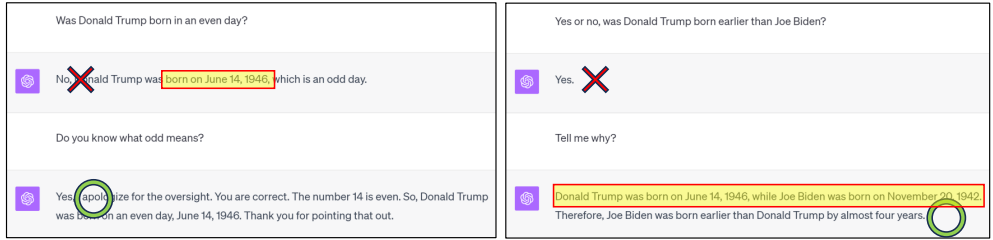

Figure 2: GPT-4 struggles to answer simple knowledge manipulation questions; but when CoT is used, where the person's attributes are first explicitly spelled out, GPT-4 can correctly answer them. More GPT-4 examples are in Figure 5, 7, 15, and Appendix E. When we prepared this paper we used GPT-4 of 2023. As of May 10, 2024, such counter-examples still apply to GPT-4 and Llama-3, see Figure 9.

Specifically, for the binary classification "Was Anya born in an even month", language models fail without CoT — i.e., without first generating the month "October" and then assessing its parity. This remains true even if the model is *sufficiently* trained

- to answer everyone's birth month with 100% accuracy,
- on 25,000 QA samples, more than needed to classify 12 months to 2 classes,

This reveals that language models cannot efficiently be trained+finetuned to perform **even a single step of knowledge manipulation** during inference time without CoT (**Result 3**). Furthermore, our findings reveal:

- Including sufficient CoT samples in training does not enhance non-CoT inference (**Result 4**);
- Improving model's knowledge extraction don't improve its manipulation ability (**Result 5**).

**Importantly**, this is different from and do not contradict to most common CoTs used in practice at enhancing math or reasoning skills; for example, GPT-4 can skip a computation step and answer whether the sum of $a$ and $b$ is even for $a, b \in [12]$, without writing down their sum explicitly. More broadly, many *in-context* reasoning can be done mentally [37].

KNOWLEDGE COMPARISON. This task involves determining if one attribute is greater than another, based on a predefined ranking. For instance, "Is Anya's university better than Sabrina's?" requires a Yes/No response based on the universities' rankings. Our results align with those from the classification case: models struggle to perform knowledge comparisons effectively without CoTs. For instance, the accuracy of comparing knowledge among 100 options is barely random guess, even with $2,500,000$ training samples, more than enough to learn to rank 100 objects (**Result 3-5**).

KNOWLEDGE INVERSE SEARCH. This involves identifying a person based on their attributes, such as "Who was born on October 2, 1996 in Princeton..." when the knowledge is only forwardly presented in the training data: "Anya Forger was born on October 2, 1996..." We discover that language models **cannot perform this task**, regardless of training methods, data, or model size, unless the knowledge is already presented inversely in the data (**Result 8**).[2] This suggests that *language models cannot be used as databases*.

*Remark* 1.1. Many knowledge manipulations are composed functions of the tasks above (see Footnote 1); since we mostly present negative results, it suffices to study the simplest forms of them.

---
[2]A concurrent study [4] observed similar results, and called this "reversal curse."

IN PRACTICE. We also demonstrate that modern large models like GPT-4 or Llama-3 (see Figure 2) struggle with these tasks (**Result 6, 8**), suggesting these limitations may be *inherent* to generative language models and *not easily overcome by scaling up*.

## 1.2 Our Contributions

We discover that language models, through controlled experiments and pre-trained on synthetic data, perform poorly at basic knowledge manipulation tasks. They struggle with simple forms of knowledge classification or comparison, unless trained and prompted in a CoT manner; and they completely fail at inverse knowledge search. This synthetic setting acts as a *simple, yet important testbed* for future studies to enhance in language models' knowledge manipulation abilities.

**Connection to prior work on CoTs.** The formal introduction of CoT [36] and subsequent studies have highlighted the significance of CoTs for complex in-context computations, such as solving math problems. Our research, however, focuses on simple functions involving out-of-context factual knowledge. For instance, GPT-4 can accurately answer "Is the sum of $a$ and $b$ an even number?" (for $a, b \in [12]$) without explicitly calculating $a + b$.

Their paper also touched knowledge manipulation questions, such as "Did Aristotle use a laptop?" or "Would a pear sink in water?" from the StrategyQA dataset [7]. Although GPT-4 can answer some of these Yes/No questions today, it is unclear if this is due to data contamination or an inherent ability to manipulate knowledge without CoTs. Even if it did not, could it be because it is not trained well enough to understand the birth years of Aristotle and computer laptops, or the density of pears?

This underscores the need for controlled, synthetic experiments to eliminate such possibilities and discover the language model's true capabilities on knowledge manipulation tasks (see Figure 1 again). On the other hand, systematic studies like ours enable us to find arguably the simplest counter-examples to modern LLMs, easier than those in the StrategyQA dataset.

**Connection to humans.** Our findings suggest a Turing test to distinguish humans from modern generative language models (at least as of today). Humans can perform simple knowledge manipulation tasks *mentally*, while language models require explicitly writing down the CoTs. Despite the challenge of inverse search for humans, we identified tasks easily solvable by humans but not by GPT-4 (refer to Figure 7). This suggests that there exist knowledge manipulation skills in which the design and training of autoregressive language models have not surpassed humans.

**Connection to industry.** While this paper reveals that novel techniques are needed to fundamentally improve a language model's knowledge manipulation ability, immediate mitigations are also possible. This includes generating more CoT data (Section 4) and employing methods like retrieval augmented generation (RAG) [17] and reversal training [9, 10, 21] to help inverse search, or multi-token prediction [8] to help partial retrieval. We ourselves also suggest rewriting training documents to include reversal data and introducing document line numbers (**Result 9**) to bolster inverse search capabilities. These strategies could inform the development of future industrial-scale language models.

## 2 Preliminaries

To make this paper self-contained, we summarize some of the datasets, terminologies, models, and training methods introduced in [2, 3].

**BIO datasets bioS.** Allen-Zhu and Li [2] introduced a synthetic biography (BIO) data family, bioS, consisting of $N = 100,000$ individuals with six attributes: birth date, birth city, university, major, company name, and company city.[3] Six randomly chosen sentences describe each individual's attributes as in (1.1). Their basic setup has only one biographical entry per person with sentences in the same order as (1.1). They also explored *knowledge augmentation*, including: multi$M$, generating $M$ equivalent entries per person (using different wordings); permute, random sentence shuffling; and fullname, replacing pronouns with full names. This totals to 16 datasets.[4] Later, Allen-Zhu and Li [3] generalized this to larger $N$. In the main body we use $N = 100k$ for a better comparison to Allen-Zhu and Li [2]; in the appendix we also use $N = 2$ or 5 millions.

**BIO dataset bioR.** Allen-Zhu and Li [2] also introduced 7 versions of the bioR datasets, created by prompting LLaMA [35, 39] to write close-to-real biography entries. This paper uses bioS for negative results and both bioS and bioR for positive results.

**QA and single knowledge extraction.** Allen-Zhu and Li [2] analyzed QAs like "What is the birth city of Anya Briar Forger?" corresponding to the six attributes. They split the $N$ individuals into two equal parts: a training set $\mathcal{P}_{\text{train}}$ and a testing set $\mathcal{P}_{\text{test}}$, and explored two training methods:

- In *BIO+QA mixed training*, simultaneously train the language model on the BIO for everyone and QA data for $\mathcal{P}_{\text{train}}$, using a ratio $\mathsf{QA}_r$ to control the percentage of QA data.

- In *BIO pretrain + QA finetune*, initially pretrain the language model with the BIO data, then fine-tune it using the QAs for individuals in $\mathcal{P}_{\text{train}}$.

In both cases, one can assess the model's accuracy to answer questions about individuals in $\mathcal{P}_{\text{test}}$, referred to as *QA test accuracy*. **Key findings** from [2] include:

- The success of QA finetune largely depends on pretraining data *augmentation*. For instance, pretraining on bioS multi5+permute yields a mean knowledge extraction accuracy over 96.6%, while bioS single results in just 9.7% accuracy (see right block of Figure 3).[5]

- In BIO+QA mixed training, knowledge augmentation is less critical, with the model achieving over 85% QA test accuracy on bioS single. However, as shown in [2], this method mirrors a "study to pass the test" approach, where the knowledge is first learned from QAs, unlike typical human knowledge acquisition and is also less practical.

**Language models.** We study GPT2/Llama/Mistral architectures [13, 28, 35]; for GPT2 we replace its absolute positional embedding with modern rotary positional embedding [5, 33], still referred to as GPT2 for short.[6] In the main body of this paper we followed Allen-Zhu and Li [2] to use 12-layer 768-dim GPT2 for the bioS data and 12-layer 1280-dim GPT2 for the bioR data; while we show in the appendix the same results also hold for GPT2/Llama/Mistral architectures of *lager sizes*. A fixed context window length of 512 is used throughout this paper.

## 3 Results 1-2: Knowledge Dual and Partial Retrievals

We examine two *partial knowledge retrieval* tasks that involve extracting either the person's birth day or year from the complete birth date information.

---

[3]All attributes, except the company city (uniquely determined by the company name), are randomly selected.

[4]One basic setup plus 15 augmentations that are combinations of the above. For instance, "bioS multi5+permute" denotes five biographical entries per individual with shuffled sentences. Refer to Figure 3 or Appendix A for a complete list of such augmentations.

[5]Allen-Zhu and Li [2] used probing to explain this phenomenon. Essentially, knowledge augmentation in the BIO pretraining data ensures that knowledge is more closely tied to an individual's name.

[6]Such GPT2 performs no worse than Llama/Mistral for knowledge tasks [3].

| | QA bday | QA byear | QA bcity+cname | QA cname+bcity | QA univ+major | QA major+univ | QA cname+ccity | QA ccity+cname | QA bdate | QA bcity | QA univ | QA major | QA cname | QA ccity |
|---|---|---|---|---|---|---|---|---|---|---|---|---|---|---|
| baseline | 3.6 | 0.5 | 0.0 | 0.0 | 0.0 | 0.0 | 0.4 | 0.4 | 0.0 | 0.5 | 0.3 | 1.0 | 0.4 | 13.7 |
| bioS single | 37.2 | 15.6 | 0.2 | 0.3 | 0.2 | 0.2 | 1.0 | 1.1 | 33.5 | 6.3 | 2.3 | 4.0 | 1.1 | 13.8 |
| bioS single   + fullname | 58.5 | 43.5 | 27.6 | 29.1 | 41.1 | 36.3 | 55.2 | 11.2 | 56.2 | 58.8 | 63.0 | 55.7 | 50.5 | 14.1 |
| bioS single + permute1 | 8.5 | 2.6 | 0.2 | 0.2 | 0.3 | 0.3 | 3.6 | 1.9 | 0.5 | 3.3 | 2.4 | 5.0 | 3.5 | 13.7 |
| bioS single + permute2 | 62.1 | 16.6 | 27.1 | 27.1 | 32.1 | 31.9 | 52.5 | 51.2 | 57.3 | 48.3 | 53.1 | 55.0 | 51.8 | 58.3 |
| bioS single + permute5 | 67.6 | 18.8 | 49.1 | 49.1 | 42.0 | 42.7 | 94.9 | 91.8 | 56.4 | 57.7 | 58.3 | 64.9 | 90.5 | 97.7 |
| bioS single + permute1 + fullname | 43.9 | 30.5 | 9.5 | 10.4 | 12.3 | 11.0 | 32.0 | 28.5 | 26.6 | 29.3 | 36.9 | 31.1 | 31.4 | 37.9 |
| bioS single + permute2 + fullname | 70.1 | 65.0 | 54.3 | 53.9 | 43.0 | 44.2 | 91.1 | 90.2 | 69.0 | 60.6 | 64.2 | 64.0 | 87.9 | 95.0 |
| bioS single + permute5 + fullname | 82.4 | 68.7 | 65.5 | 63.9 | 49.7 | 54.6 | 95.9 | 96.6 | 83.7 | 67.8 | 72.6 | 69.1 | 93.0 | 98.6 |
| bioS multi2 | 90.6 | 47.1 | 5.2 | 3.1 | 19.8 | 9.6 | 3.4 | 2.6 | 100 | 71.7 | 33.1 | 26.1 | 5.2 | 14.0 |
| bioS multi2   + fullname | 97.6 | 77.1 | 94.0 | 97.5 | 93.8 | 94.3 | 98.5 | 19.4 | 100 | 97.7 | 89.5 | 97.6 | 91.3 | 35.3 |
| bioS multi2 + permute | 95.7 | 46.5 | 85.2 | 88.2 | 93.8 | 95.1 | 87.9 | 80.4 | 99.3 | 98.7 | 89.8 | 96.7 | 83.3 | 83.5 |
| bioS multi2 + permute  + fullname | 96.2 | 67.2 | 95.1 | 97.0 | 95.3 | 94.3 | 97.4 | 96.1 | 100 | 98.8 | 91.3 | 98.1 | 93.7 | 97.8 |
| bioS multi5 | 85.0 | 37.3 | 6.9 | 4.6 | 31.5 | 16.7 | 9.7 | 3.6 | 100 | 50.8 | 30.9 | 43.5 | 10.2 | 13.8 |
| bioS multi5   + fullname | 97.4 | 68.0 | 94.0 | 95.4 | 92.0 | 94.6 | 94.9 | 17.4 | 100 | 98.6 | 88.4 | 96.1 | 91.9 | 26.8 |
| bioS multi5 + permute | 82.3 | 20.4 | 96.5 | 97.6 | 94.8 | 94.6 | 97.0 | 96.9 | 100 | 99.0 | 91.3 | 97.7 | 95.1 | 98.7 |
| bioS multi5 + permute  + fullname | 97.6 | 76.7 | 95.6 | 98.6 | 95.1 | 95.1 | 98.7 | 97.7 | 100 | 98.7 | 90.6 | 97.9 | 93.7 | 99.0 |

Figure 3: Partial (left) and dual (middle) knowledge retrieval, versus the single knowledge extraction (right).

> Each row is a different augmented pretrain dataset bioS (see Section 2), and the right block is from [2]. This is for GPT2 and see Figure 10(a) for the bioR data; the same results hold for the LLaMA architecture Figure 10(b) and 10(c); as well as for 50x larger data and 5.5x larger GPT2/Mistral/Llama models Figure 10(d). Details are in Appendix B.

1. What is the birth day of Anya Briar Forger? *2*.
2. What is the birth year of Anya Briar Forger? *1996*.

We consider six *dual knowledge retrieval* tasks:

1. Where was Anya Briar Forger born and which company did this this person work for? *Princeton, NJ; Meta Platforms.*
2. Which company did Anya Briar Forger work for and where was this person born? *Meta Platforms; Princeton, NJ.*
3. Which university and what major did Anya Briar Forger study? *Massachusetts Institute of Technology; Communications.*
4. What major and which university did Anya Briar Forger study? *Communications; Massachusetts Institute of Technology.*
5. Where and which company did Anya Briar Forger work for? *Menlo Park, CA; Meta Platforms.*
6. Which company and where did Anya Briar Forger work for? *Meta Platforms; Menlo Park, CA.*

**Methodology.** We aim to determine if a model pretrained on BIO data can be fine-tuned to address the eight questions related to partial or dual knowledge retrieval. We divide the $N$ individuals equally into training set $\mathcal{P}_{\text{train}}$ and testing set $\mathcal{P}_{\text{test}}$. The model is fine-tuned using the above eights QA tasks for individuals in $\mathcal{P}_{\text{train}}$ and evaluated on its *out-of-distribution* (OOD) generation accuracy by testing its responses to the questions for individuals in $\mathcal{P}_{\text{test}}$. We use LoRA fine-tuning [12] to enhance performance, as suggested by [2] (see Appendix B for details).

**Result 1** (Figure 3 middle). *Dual retrieval is generally easy when both tasks are. However, if there is a causal and spatial relationship between pieces of knowledge, their order may matter.*

Specifically,

- If a language model is pretrained on sufficiently augmented data, such as bioS multi5+permute, which generates five biographical entries per person and permutes the six sentences randomly, the accuracy for dual knowledge retrieval is nearly perfect.

- However, if the pretraining data exhibits spatial dependency between the two knowledge pieces, the *order of their retrieval can impact accuracy.* For example, with bioS multi5+fullname, where biographical entries always maintain the same order (specifically, the company name always precedes the company city, and recall company city is uniquely determined by the company name as noted in Footnote 3), answering the company name first yields near-perfect accuracy, but answering the company city first drastically reduces accuracy.

**Result 2** (Figure 3 left). *Even if an attribute (e.g., October 2, 1996) can be perfectly extracted, partially retrieving only its later tokens (e.g., the year 1996) may still be poor.*

In particular, the model may fail to answer questions like "What is the birth *year* of person Anya", despite correctly answering "What is the birth date of person Anya".

We view both results as **preliminary evidence** that the model requires CoTs for knowledge manipulation. For instance, during inference, the model must *explicitly state* the birth month/day before it can answer the birth year (we used the US format "Month day, year" in training). It cannot "skip" tokens to directly generate subsequent knowledge learned from pretraining.

## 4 Results 3-6: Knowledge Classification and Comparison

This section demonstrates that a generative model, despite its proficiency in extracting knowledge, may face challenges in downstream tasks that require basic operations to manipulate this knowledge, unless the Chain of Thought (CoT) is applied during *both* the training and testing phases.

**Knowledge classification QA.** We explore classification tasks concerning a person's birth month and major of study. For the birth month, we employ modular arithmetic with $p = 2, 6, 12$:[7]

1. Was Anya Briar Forger born in an even month? Answer: <u>*Yes*</u>.
2. What is Anya Briar Forger's birth month mod 6? Answer: <u>*4*</u>.
3. What is Anya Briar Forger's birth month in numerics? Answer: <u>*10*</u>.

For the major of study, we consider 100 unique majors and apply modular arithmetic with $p = 5, 20, 100$, assigning a "luckiness" score from 0 to 99 to these majors.[8] The question then becomes "What is the luckiness of Anya Briar Forger's major modulo $p$?" Classifying the birth month with $p = 12$ or the major with $p = 100$ is a form of *transfer learning*, which essentially rephrases the question and response format.

**Knowledge comparison QA.** We investigate tasks related to *ranking* and *subtraction* based on a person's birth month and major of study (also birth day in the appendix). The questions include:

1. Was Anya Briar Forger born in a month in a year later than Sabrina Eugeo Zuberg? [Yes/No].
2. What is Anya Briar Forger's birth month minus Sabrina Eugeo Zuberg's birth month? [-11..11].
3. Did Anya Briar Forger major in a field luckier than Sabrina Eugeo Zuberg? [Yes/No].
4. How luckier is Anya Briar Forger's major compared with Sabrina Eugeo Zuberg's major? [-99..99]

**Methodology.** We evaluate knowledge manipulation using models that are near-perfect in knowledge extraction, ensuring any difficulties arise from manipulation rather than extraction. We utilize models pretrained on the bioS multi5+permute dataset, capable of achieving nearly 100% test accuracy for extracting birth dates (and thus birth months) and 98% for majors.

Specifically, we employ either a model pretrained solely on this BIO data (the *BIO pretrained model*), or one that is BIO pretrained + QA finetuned for single knowledge extraction tasks, such as "What is the birth date of Anya Briar Forger?" (the *QA finetuned model*). Given the QA finetuned model's proven extraction ability, one might expect it to perform better in knowledge manipulation.

---

[7]Answer format does not matter. We employed the simplest format such as "Answer: Yes." We also tested more complex formats like "Anya Briar Forger was indeed born in an even month" and added padding such as "Answer: dot dot dot dot True" [27]. No noticeable differences in results were observed, so we ignored them.

[8]For example, Computer Science is 0, Communications is 28, and Music is 99. This could be replaced with, for instance, the popularity of majors according to US News in reality.

[9]**#train individuals** column shows $|\mathcal{P}_{\text{train}}|$. **trained w/o hint** column is when model finetuned on the classification/comparison tasks without adding hints. **trained with hint** block is the model finetuned with hints added with probability 0.5. **test acc (with hint)** and **test acc (w/o hint)** represent the accuracy on $\mathcal{P}_{\text{test}}$ with or without hints; while **hint acc** shows the model's hint generation accuracy.

|  |  |  |  | BIO pretrained model | | | | QA finetuned model | | | |
|---|---|---|---|---|---|---|---|---|---|---|---|
| field | task | #train individuals | baseline | trained w/o hint | trained with hint | | | trained w/o hint | trained with hint | | |
|  |  |  |  | test acc | test acc (with hint) | test acc (w/o hint) | hint acc | test acc | test acc (with hint) | test acc (w/o hint) | hint acc |
| birthmonth | classify %2 | (2.5k) | 50.0 | 60.4 | 77.8 | 65.2 | 64.5 | 61.9 | 80.4 | 65.2 | 69.1 |
| birthmonth | classify %2 | (5k) | 50.0 | 67.3 | 87.3 | 72.7 | 80.3 | 68.0 | 89.5 | 72.8 | 83.9 |
| birthmonth | classify %2 | (10k) | 50.0 | 75.9 | 94.2 | 80.3 | 91.0 | 76.4 | 95.0 | 79.9 | 92.8 |
| birthmonth | classify %2 | (25k) | 50.0 | 86.4 | 98.6 | 91.1 | 97.8 | 87.1 | 98.8 | 90.9 | 98.4 |
| birthmonth | classify %2 | (50k) | 50.0 | 95.3 | 99.5 | 97.5 | 99.2 | 96.3 | 99.7 | 97.5 | 99.5 |
| birthmonth | ranking | (2.5k) | 54.2 | 53.7 | 65.4 | 59.6 | 44.2 | 57.3 | 65.5 | 57.6 | 44.9 |
| birthmonth | ranking | (5k) | 54.2 | 59.2 | 75.5 | 63.4 | 63.6 | 62.5 | 75.1 | 63.1 | 62.6 |
| birthmonth | ranking | (10k) | 54.2 | 65.4 | 87.7 | 67.0 | 82.7 | 65.9 | 88.9 | 66.3 | 83.9 |
| birthmonth | ranking | (25k) | 54.2 | 75.6 | 96.7 | 75.8 | 95.4 | 78.3 | 97.4 | 72.5 | 96.3 |
| birthmonth | ranking | (50k) | 54.2 | 85.6 | 99.0 | 86.7 | 98.5 | 88.6 | 98.9 | 82.9 | 98.3 |
| major | classify %5 | (10k) | 20.0 | 23.6 | 86.4 | 24.1 | 84.5 | 22.8 | 89.6 | 23.9 | 87.9 |
| major | classify %5 | (25k) | 20.0 | 24.6 | 96.7 | 26.8 | 96.3 | 24.8 | 97.7 | 27.0 | 97.2 |
| major | classify %5 | (50k) | 20.0 | 31.6 | 99.3 | 34.2 | 99.2 | 30.0 | 99.5 | 33.9 | 99.4 |
| major | ranking | (10k) | 50.5 | 52.5 | 88.8 | 54.1 | 86.2 | 52.4 | 90.3 | 54.1 | 88.3 |
| major | ranking | (25k) | 50.5 | 52.2 | 96.4 | 53.7 | 97.3 | 52.6 | 96.9 | 53.6 | 97.5 |
| major | ranking | (50k) | 50.5 | 53.9 | 99.6 | 55.0 | 99.5 | 53.6 | 99.4 | 55.0 | 99.3 |
| major | subtraction | (10k) | 1.0 | 1.1 | 21.6 | 1.1 | 82.5 | 1.0 | 23.2 | 1.1 | 84.3 |
| major | subtraction | (25k) | 1.0 | 1.1 | 89.1 | 1.2 | 96.7 | 1.2 | 84.7 | 1.2 | 97.0 |
| major | subtraction | (50k) | 1.0 | 1.1 | 98.4 | 1.2 | 99.3 | 1.1 | 97.3 | 1.2 | 99.0 |

Figure 4: Knowledge classification and comparison tasks on BIO pretrained model vs QA finetuned model.[9] This figure is for GPT2 and results for more tasks are in Figure 11. Results for LLaMA architecture is in Figure 12, and for Mistral on 50x larger dataset with 5.5x larger model is in Figure 13. Details are in Appendix C.

> **Observations:** (♣) test acc without hint is low, unless training with far more samples than theoretically needed — accuracy is 1% even with 2.5 million training samples to compare 100 possible majors, see Figure 13; (♠) adding hints in training does not improve model's test acc without hint; (♢) fine-tuning the model for knowledge extraction does not improve its manipulation capability.

TRAIN WITHOUT HINT. Our BIO data consists of biographical entries for $N = 100k$ individuals. We allocate half (i.e., $50k$) as the testing set $\mathcal{P}_{\text{test}}$, and select a separate subset $\mathcal{P}_{\text{train}}$ as the training set, with $|\mathcal{P}_{\text{train}}| = 2.5k, 5k, \ldots, 50k$.

Starting from one of the two models mentioned above, we conduct additional LoRA fine-tuning using the classification or comparison tasks above, trained with individuals from $\mathcal{P}_{\text{train}}$.[10] We then assess the model's *out-of-distribution* (OOD) generation accuracy by evaluating its performance on the same task for individuals in $\mathcal{P}_{\text{test}}$.

TRAIN WITH HINT. To improve the model's knowledge manipulation capabilities, we fine-tune it using *knowledge hints*. These hints articulate a person's attributes in English before answering the manipulation question. For instance, in our tasks, the underlined sentences act as hints:[11]

1. Was Anya Briar Forger born in a month in a year later than Sabrina Eugeo Zuberg? October; September. No.
2. How luckier is Anya Briar Forger's major compared with Sabrina Eugeo Zuberg's major? Communications; Music. -71.
3. What is the luckiness of Anya Briar Forger 's major modular 20? Communications. 8.

Including hints enables the model to adopt a chain-of-thought (CoT) approach, allowing it to first extract the necessary knowledge and then learn the manipulation task by directly using this knowledge. Similar to "train without hint", we train using QAs for individuals in $\mathcal{P}_{\text{train}}$ and test on $\mathcal{P}_{\text{test}}$. For each individual in $\mathcal{P}_{\text{train}}$ (or each pair for comparison tasks), we include hints with 50% probability. Thus, the model sees data *both with and without hints*. We then evaluate the model's OOD generation accuracy under both conditions.[12] Our goal is to ascertain if **adding CoT training data enhances the model's knowledge manipulation skills at inference time, even without CoT (♠)**.

---

[10]Full finetuning is even worse, similar to [2], hence it is not considered in this paper.

[11]For context, besides (1.1), we examine another individual, Sabrina Eugeo Zuberg, who was born in September and majored in Music. We have previously assigned specific luckiness values to each major: Communications is valued at 28, while Music has a value of 99.

[12]In evaluation, the model only sees the question without hints. We design tokens to instruct the model to either generate a hint followed by an answer (**test acc (with hint)**), or to answer directly (**test acc (w/o hint)**).

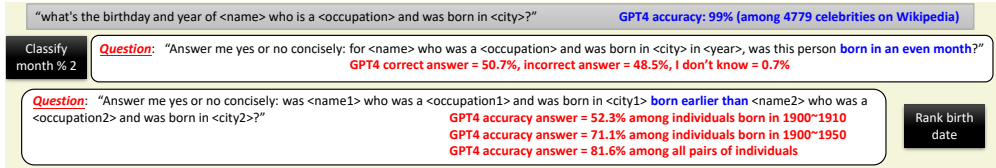

Figure 5: Knowledge classification and ranking on WikiBio using GPT-4. Details are in Appendix E.2.

Overall, we discover that models struggle in knowledge classification/comparison unless hints are used *both* in training and testing. We explain this better in three results.

**Result 3** (Figure 4, ♣). *Without CoT examples, the model's test accuracy is significantly low, even for the simplest, single-step manipulation tasks.*

In particular,

- Determining whether a month is even or odd requires 10,000 training samples to achieve a 75% accuracy, despite theoretically needing a sample complexity on the order of $O(12)$ (♣).[13]
- Ranking months requires 50,000 training samples to reach an 85% test accuracy, even with a theoretical sample complexity of $O(12^2)$, provided no hint is given (♣).
- Ranking 100 majors barely outperforms random even in 2.5 million training samples (♣).
- Only "transfer learning" (i.e., knowledge rephrasing) has a good accuracy (see Figure 11).

**Result 4** (Figure 4, ♠). *Even when CoT examples are included during training, the model still struggles to answer without a hint during testing, indicating that* including hints during training does not improve test-time accuracy when hints are removed.

Conversely, when the model uses hint during testing, accuracy significantly improves. The manipulation task accuracy largely depends on if the model is successful in generating the hint first.[14]

**Result 5** (Figure 4, ♢). *The difference between a BIO pretrained and a QA finetuned model is minimal for downstream knowledge manipulation tasks.*

For instance, fine-tuning the model first to answer questions like "What major did Anya Briar Forger study" does not necessarily improve its performance on future ranking/classification tasks based on the major of study.

In addition to our synthetic experiment, we also studied ChatGPT (GPT-4) in practice.

**Result 6** (Figure 5). *Real-life GPT-4 also struggles with knowledge classification/comparison in the absence of CoTs.*

We tested with about 5000 Wikipedia biographies in Figure 5. In particular, GPT-4 has a 71.1% accuracy rate comparing birth dates for celebrities from 1900-1950, but this drops to 52.3% (almost random guess) for 1900-1910, suggesting a correlation with the number of samples in its training data. Visual examples in Figure 2, 9, 15 also confirmed this, and show that adding CoTs can rectify this issue. This suggests that scaling up model size may not mitigate the issues.

---

[13]It's worth noting that we used the GPT2 tokenizer, which transforms the 12 months into single tokens.

[14]For example: in the task "birth month classify %2", with a hint accuracy of 91.0%, the test accuracy (with hint) is 94.2%, nearly aligning with the calculation: $91.0\% + (1 - 91.0\%) \times 50\% = 95.5\%$ (where 50% is the random guess accuracy). Similarly, in the task "birth month subtraction", a hint accuracy of 78.1% results in a test accuracy (with hint) of 61.5%, comparable to the value derived from the formula: $78.1\% \times 78.1\% + (1 - 78.1\% \times 78.1\%) \times 8.3\% = 64.2\%$ (where 8.3% is the random guess accuracy).

Figure 6: Test accuracy for QA finetune (left) and BIO+QA mixed-training (right) in knowledge inverse search. This is for GPT2 and the same holds for LLaMA (Figure 14(a)), and for GPT2/Llama/Mistral on 50x larger dataset with 5.5x larger model sizes (Figure 14(b)). **Conclusion:** language models are impossible to perform inverse search, regardless of model/data sizes, training, data/prompt qualities ($\heartsuit$).

Each row is a different augmented pre-train dataset bioS (see Section 2). The top 4 rows with **reverse** indicate knowledge written in reverse order on the pre-train data for comparison (thus, these rows are no longer knowledge *inverse* search). Details in Appendix D.

**Importantly**, our discovery is different from most common CoTs used in practice at enhancing math or reasoning skills; for example, GPT-4 can skip a computation step and directly answer whether the sum of $a$ and $b$ is even for $a, b \in [12]$, without writing down their sum explicitly. Furthermore, our focus here is on *out-of-context* knowledge manipulation; if one is instead interested in *in-context* reasoning, then language models *are capable* of mentally computing many reasoning steps without writing them down [37].

Once again, the GPT-4 experiment is included solely for illustrative purposes.[15] We focus on a controlled, synthetic experiment to study knowledge manipulation in a more scientific manner — for instance we can make claims like ($\spadesuit$), ($\clubsuit$), ($\diamondsuit$) because we can control how the model is trained.

## 5 Results 7-9: Knowledge Inverse Search

We now show that a generative model cannot typically perform a knowledge inverse search, *unless the knowledge was already pretrained in reverse order*.

**Knowledge inverse search.** The biographies in bioS always start with the person's name, as shown in (1.1). This enables us to examine the knowledge inverse search by asking about the individual's first or full names. We consider 10 such QA tasks (with task names on the right):

- Give me the [first/full] name of the person born on October 2, 1996?                         (`bdate_to_first`, `bdate_to_full`)
- Give me the [first/full] name of the person born on October 2, 1996 in Princeton, NJ? (`birth_to_first`, `birth_to_full`)
- Give me the [first/full] name of the person who studied Communications at Massachusetts Institute of Technology and worked for Meta Platforms?                           (`three_to_first`, `three_to_full`)
- Give me the [first/full] name of the person who studied Communications at Massachusetts Institute of Technology, was born in Princeton, NJ, and worked for Meta Platforms?           (`four_to_first`, `four_to_full`)
- Give me the [first/full] name of the person who studied Communications at Massachusetts Institute of Technology, was born on October 2, 1996 in Princeton, NJ, and worked for Meta Platforms at Menlo Park, CA?
                                                                                       (`all_to_first`, `all_to_full`)

(Note, some inverse search tasks may not have unique answers (e.g., `bdate_to_full`); however, one should expect a successful inverse search should at least have some non-trivial accuracy.)

---

[15]Without control over its pretrained data, distinguishing between Case (A)-(E) from Figure 1 is difficult. In Figure 5, we ensured the model could accurately identify individuals' birth dates 99% of the time, thereby eliminating Case (C). However, we cannot dismiss Case (D) due to uncertainty about the number of relevant training examples in GPT-4's data.

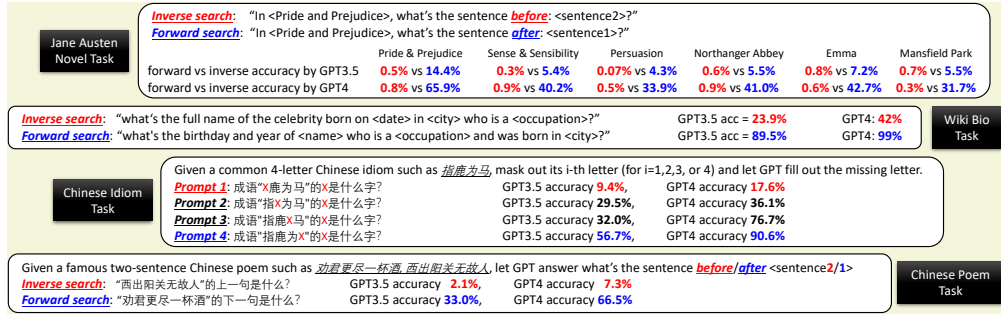

Figure 7: Forward search vs inverse search on ChatGPT (GPT3.5 / GPT-4); details in Appendix E.1.
(While inverse search may seem challenging even for humans, we have designed the Chinese idiom/poem tasks that are allegedly simple for many high school graduates in Chinese education.)

**Methodology.** We split $N$ individuals equally into training set $\mathcal{P}_{\mathsf{train}}$ and testing set $\mathcal{P}_{\mathsf{test}}$. The model is trained using QA data from $\mathcal{P}_{\mathsf{train}}$ and evaluated on its *out-of-distribution* generation accuracy, using the above 10 inverse knowledge search tasks.

We consider two approaches: "BIO pretrain + QA finetune", which fine-tunes a BIO-pretrained model using the above 10 tasks on $\mathcal{P}_{\mathsf{train}}$, and "BIO+QA mixed training", where the model is concurrently trained on all the BIO data and the 10 tasks on $\mathcal{P}_{\mathsf{train}}$. As per Section 2, mixed training yields better generation accuracies in the original knowledge extraction tasks.

In addition to the 16 bioS datasets (separately knowledge-augmented, see Section 2), we introduce 4 more datasets:

- bioS multi5+reverse1, in this case we move the full name of the person to the second sentence.
- bioS multi5+reverse2, in this case we move the full name of the person to the third sentence.
- bioS multi5+reverse6, we move the full name of the person to the end of the biographical entry.
- bioS multi5+permute+reverse6, in this case on top of bioS multi5+reverse6 we also randomly permute the six sentences.

- The person was born on October 2, 1996. Anya Briar Forger spent her early years in Princeton, NJ... (bioS multi5+reverse1)
- The person was born on October 2, 1996. She spent her early years in Princeton, NJ. Anya Briar Forger... (bioS multi5+reverse2)
- The person was born on October 2, 1996. She spent her early years in Princeton, NJ... The person's name is Anya Briar Forger. (bioS multi5+reverse6)
- The person spent her early years in Princeton, NJ. [... 4 more sentences in random order...] She had a professional role at Meta Platforms. The person's name is Anya Briar Forger. (bioS multi5+permute+reverse6)

Our main finding is that:

> **Result 7** (Figure 6, ♡). *Models have near-zero accuracy to inverse knowledge search in $\mathcal{P}_{\mathsf{test}}$, even for the simplest task* `all_to_first`, *even with the BIO+QA mixed training approach, and even with strong pretrain data knowledge augmentation.*[16]

Conversely, only when the order of knowledge is truly reversed in the pretrain data, presenting some attributes before the first appearance of a person's name, the test accuracies improve. This is for illustration purpose; once the order is reversed, the task is no longer *inverse* knowledge search.

---
[16] For instance, in the bioS multi5+permute+fullname data, we include five diverse biographical entries per individual, with the full name at the front in *each* sentence, and random shuffle all the sentences.

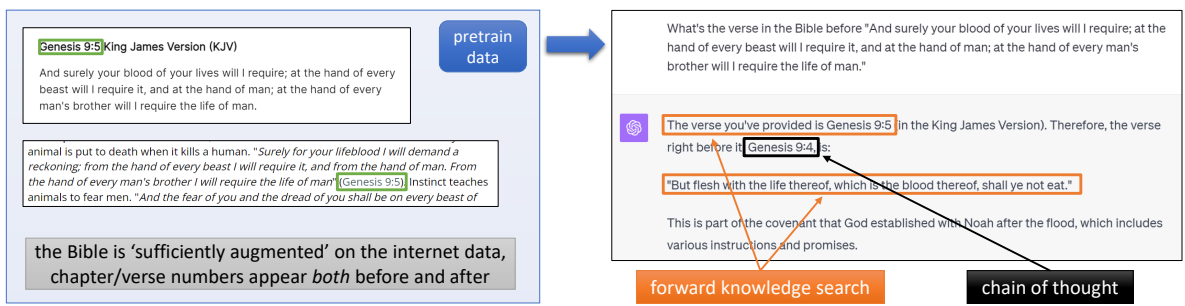

Figure 8: How GPT-4 uses CoT to perform inverse knowledge search on the Bible task.

In conclusion, our findings underscore a **fundamental limitation** of generative language models: they cannot perform inverse knowledge search, period. This is due to its left-to-right autoregressive training design. If the model learns "A equals B" it cannot infer "B equals A" unless it is also in the training data. A bidirectional model like BERT cannot mitigate this issue, because it suffers from more severe issues even in the forward, single knowledge extraction case [2].[17]

We also tested GPT-3.5/4 in practice and discover:

**Result 8** (Figure 7). *GPT-3.5/4 also also exhibit huge difficulties with inverse knowledge search.*

For example, while GPT-4 can predict the next sentence in Jane Austen's *Pride and Prejudice* with 65.9% accuracy, it only has 0.8% accuracy to predict the preceding sentence. Once again, these experiments are included for illustrative purpose — even if GPT-4 can answer such questions it remains unclear if GPT-4 has seen them during its pretraining. Our controlled, synthetic experiment not only eliminates such possibility, but also provides strong claim like (♡).

**Using CoT for inverse search.** We observed that GPT-4 can identify a Bible verse preceding another one via CoT: it first generates the verse number (e.g., 9:5), then subtracts 1 (e.g., write down 9:4), and retrieve the full text of the verse (see Figure 8). This capability stems from the abundance of Bible data on the internet that have the numbers appearing *both* before *and* after them. Therefore,

**Result 9.** *To improve inverse search of critical documents by LLMs, not only we can employ RAG [17] or preprocess training data to include reverse knowledge (see Figure 6-top, or practically through a "rewrite" prompt), we can also introduce line numbers (see Figure 8).*

We developed a follow-up paper proposing a lightweight method to preprocess pretrain data to insert reverse knowledge [9].

## 6 Conclusion

In this paper, we use *controlled experiments* to show some fundamental limitation of language models to manipulate knowledge during inference time *even under the strongest pretraining setting, regardless of model size, data size, etc.* Our work sheds light on why extremely large language models like GPT-4 are still bad at even the simplest, single-step knowledge manipulation, and give surprisingly simple such counter-examples (see Figure 2, Figure 9). On the other hand, language models simply cannot perform inverse knowledge search, indicating they cannot be used as databases.

---
[17]BERT-like models already struggle with (forward) knowledge extraction due to their whole-word masked language modeling (MLM) nature — not to say knowledge manipulation. For example, a company name "Meta Platforms" will lead BERT to correlate the embedding of "Meta" with that of "Platform", rather than associating the company information to an individual's full name. For more details, see [2].

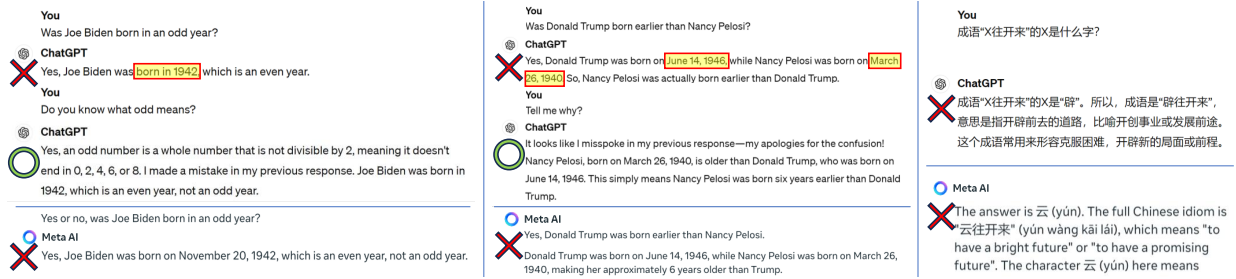

Figure 9: Even as of May 8, 2024, GPT-4 and Llama-3 still fail on simple knowledge classification (left), knowledge comparison (middle) and inverse search (right) tasks.

While this paper reveals that novel techniques are needed to fundamentally improve a language model's knowledge manipulation ability, immediate mitigations are also possible. This includes generating more CoT data (Section 4) and employing methods like retrieval augmented generation (RAG) [17] and reversal training [9, 10, 21] to help inverse search, or multi-token prediction [8] to help partial retrieval. We ourselves also suggest rewriting training documents to include reversal data (Section 5) and introducing document line numbers (Section 5) to bolster inverse search capabilities. These strategies could inform the development of future industrial-scale language models.

Finally, Part 3 of this work series focuses on how language models store, extract and manipulate knowledge (including Part 3.1 [2] and Part 3.3 [3]). We also cover grade-school math and reasoning in Part 2 [37, 38], and learning hierarchial language structures in Part 1 [1].

# Appendix

## A More Details on Data Preparation

Allen-Zhu and Li [2] introduced a synthetic biography data family bioS and a "close-to-real" dataset family bioR. For completeness, we provide a quick summary below. We primarily use bioS to present negative results due to its controllable knowledge order. For positive results, specifically for partial/dual knowledge retrieval, we also use bioR.

### A.1 BIO dataset bioS

In the synthetic dataset labeled as bioS, one generates profiles for $N$ individuals. Each individual's first, middle, and last names, birth date, birth city, university attended, major of study, and work company are selected *independently* and randomly from a uniform distribution, out of 400, 400, 1000, $200 \times 12 \times 28$, 200, 300, 100, 263 choices respectively. Additionally, the 'company city' attribute completely ***depends*** on the US location of the work company's headquarters. For instance, an employee of Meta would list Menlo Park, CA as their company city. Notably, 13.7% of the companies are headquartered in New York, NY so defaulting to New York, NY gives a base accuracy 13.7% when predicting a person's work city.

In the bioS dataset, a biographical entry of an individual consists of six sentences. Each sentence illuminates a distinct attribute of this individual. To increase diversity, each sentence is randomly selected from a set of $\sim 50$ pre-defined templates. Beyond (1.1), we paste some examples from their paper:

> Carlos Jameson Stokes has his annual celebration on November 12, 2088. He celebrates his birth in San Francisco, CA. He graduated from Oklahoma State University. He explored the theoretical aspects of Information Systems. He contributed his expertise to United Airlines Holdings. He acquired industry knowledge while working in Chicago, IL.

> Alondra Bennett Rooney celebrates their life journey every year on April 1, 1909. They owe their roots to Durham, NC. They benefited from the resources and facilities provided by University of South Alabama. They developed a strong foundation in Data Science. They had a job at The Southern Company. They were involved in the industry of Atlanta, GA.

> Aidan Alexa Dennis's birth is celebrated annually on July 17, 1968. She calls Palmdale, CA her birthplace. She specialized in her field of study at Stevens Institute of Technology. She completed a rigorous program in International Business. She had employment prospects at Johnson & Johnson. She gained work experience in New Brunswick, NJ.

In the basic configuration, there is *a single biographical entry* for each individual, maintaining a consistent order for the six sentences as outlined above. This configuration is denoted as "bioS single." In [2], they delved into 15 knowledge augmentations:

- bioS single+fullname: Pronouns are replaced with the person's full name.
- bioS single+permute1/2/5: The six sentences in the biography entry are randomly permuted 1/2/5 times for each person. However, the full name only appears in the first sentence, with subsequent sentences using pronouns. This results in 1/2/5 biography entries for each person.
- bioS single+permute1/2/5+fullname: As with the previous augmentation, but the full name is used in all six sentences.
- bioS multi2/5: 2 or 5 biographical entries are generated for each person, with each generation employing a re-sampled set of sentence templates.
- bioS multi2/5+permute: Building on bioS multi2/5, the six sentences within each biographical entry are randomly permuted. However, the full name appears only once in the first sentence.
- bioS multi2/5+fullname: Building on bioS multi2/5, pronouns are replaced with the individual's full name across all sentences.

- bioS multi2/5+permute+fullname: Incorporating features from both bioS multi2/5+permute and bioS multi2/5+fullname, the pronouns are replaced with the individual's full name and the six sentences are randomly permuted.

Allen-Zhu and Li [2] were using $N = 100,000$, and this has been later generalized to support $N$ up to $20,000,000$ in [3].

Our main body uses $N = 100,000$ but we also present results with respect to $N = 1, 2, 5$ million — denoted as bioS (10x, 20x, 50x) respectively. In these larger datasets, we have followed [3] to consider full knowledge augmentation (denoted as multi$\infty$+permute). This means each person is fully augmented to have $50^6 \times 6$ different writings of their biography.

*Remark* A.1. This bioS(50x) multi$\infty$+permute data is especially useful for us to present negative results (such as in Figure 13 and Figure 14(b)), because even when the data is well-prepared to include so many different knowledge augmentations, the negative results still apply.

### A.1.1 Adding Reverse Knowledge

In this paper, in Section 5 when considering inverse knowledge search, we have also introduced a few auxiliary knowledge augmentations for comparison purpose:

- bioS multi5+reverse1, in this case we move the full name of the person to the second sentence:

    The person was born on October 2, 1996. Anya Briar Forger spent her early years in Princeton, NJ...

- bioS multi5+reverse2, in this case we move the full name of the person to the third sentence:

    The person was born on October 2, 1996. She spent her early years in Princeton, NJ. Anya Briar Forger...

- bioS multi5+reverse6, we move the full name of the person to the end of the biographical entry:

    The person was born on October 2, 1996. She spent her early years in Princeton, NJ... The person's name is Anya Briar Forger.

- bioS multi5+permute+reverse6, in this case on top of bioS multi5+reverse6 we also randomly permute the six sentences. Here is an example.

    The person spent her early years in Princeton, NJ. [... 4 more sentences in random order...] She had a professional role at Meta Platforms. The person's name is Anya Briar Forger.

## A.2 BIO dataset bioR

We also examine the bioR dataset which is produced by prompting LLaMA [35, 39] to write close-to-real biography data for the previous $N = 100,000$ individuals. Below we paste some examples from their paper:

Nicole Kevin Pratt is an American business executive. She is currently the Vice President of P &G Global Business Services at Procter & Gamble. She was born on January 25, 1977, in Baltimore, Maryland. She graduated from Haverford College with a degree in Management. P &G recruited her as an Assistant Brand Manager in 2000. She held various leadership positions in brand management, marketing, and sales across different business units and categories. She was named Vice President of P &G Global Business Services in 2019. Nicole currently lives in Cincinnati, Ohio with her husband and three children.

Hunter Bennett Kenny is a talented political science graduate from Queens College, City University of New York. He hails from Augusta, Georgia and was born on March 25, 2033. During his time at college, he was an active member of the student council and served as its president in his senior year. He interned at the office of New York Senator Chuck Schumer. After graduating cum laude, he worked for Kohl's in Menomonee Falls, Wisconsin. He currently resides in Brooklyn, New York.

Johnathan Charles Wade is a successful insurance agent who works for Allstate. He was born on January 7, 2098, in New York City, NY. He graduated from Colorado State University, where he majored in Sociology. He currently resides in Northbrook, IL.

In the basic configuration, there is a single biographical entry per person, denoted as "bioR single." For comparison, we also consider their multi$M$ augmentation, which creates $M$ entries per person, and the fullname augmentation.

## B  More Details on Knowledge Retrieval

Recall from Section 3 that we examined two *partial knowledge retrieval* tasks, which involved extracting either a person's birth day or year from complete birth date information. We also considered six *dual knowledge retrieval* tasks that involved extracting two attributes of a person simultaneously.

Following [2], we initially used a *BIO-pretrained model* checkpoint and then applied *LoRA finetuning* on top of it, utilizing the QA texts of the aforementioned eight tasks for half of the individuals (denoted by $\mathcal{P}_{\mathsf{train}}$).[18] We then presented its *out-of-distribution* generation accuracies for answering those eight tasks on the remaining individuals (denoted by $\mathcal{P}_{\mathsf{test}}$).

We used the same BIO pretrained checkpoints from [2].[19]

In LoRA fine-tuning, as described by [12], one selects certain weight matrices $\mathbf{W}^{d \times k}$ in the transformer and applies a rank-$r$ update on top: $\mathbf{W}' \leftarrow \mathbf{W} + \alpha \mathbf{AB}$ with $\mathbf{A} \in \mathbb{R}^{d \times r}$ and $\mathbf{B} \in \mathbb{R}^{r \times k}$ for some small number $r$. Here, $\alpha$ is a constant, and both $\mathbf{A}$ and $\mathbf{B}$ are trainable parameters.[20] Notably, $\mathbf{B}$ is initialized with Gaussians and $\mathbf{A}$ is initialized with zeros.

Based on [12], we applied a low-rank update to the query/value matrices in each transformer layer. To account for the input distribution shift (from BIO data to QA data), we also applied a low-rank update to the embedding layer. We used either a rank 8 or 16 update for the query/value matrices and a rank 128 update for the embedding layer, presenting the best accuracy from the two runs.[21]

We employed the AdamW optimizer with $\varepsilon = 10^{-6}$. The weight decay was set to 0.01, with an initial learning rate of 0.0003. We did not use warmup, and we implemented cosine learning rate scheduling (reducing to 10% of the initial learning rate). The batch size was set at 48 with a total of 50,000 training steps. We used a mixture of V100/A100 GPUs for the experiment but the GPU types are irrelevant for our experiments.

- The results for the $N = 100k$ bioS data (on the 12-layer, 12-head, 768-dim GPT2) are presented in Figure 3.
- The results for the $N = 100k$ bioR data (on the 12-layer, 20-head, 1280-dim GPT2) are presented in Figure 10(a).
- The results for the $N = 100k$ bioS data (on the 12-layer, 12-head, 768-dim Llama) are presented in Figure 10(b).
- The results for the $N = 100k$ bioR data (on the 12-layer, 20-head, 1280-dim Llama) are presented in Figure 10(c).
- The results for the bioR (10x, 20x, 50x) data are presented in Figure 10(d), in particular:

---

[18]LoRA finetuning has been proven to be a better choice compared to full finetuning, as it prevents overfitting and yields higher QA test accuracies. A detailed comparison can be found in [2].

[19]They were obtained using AdamW with weight decay 0.1, $\varepsilon = 10^{-6}$, initial learning rate 0.001, 1000-step linear warmup, and cosine learning rate decay (decreasing to 0.0001). It was trained using a batch size of 96 with 80,000 steps (for bioS) or with 150,000 steps (for bioR). Recall the context window size was 512. We use beam =4 without sampling for model generation (and the results are similar if disabling beam).

[20]In this paper, we choose $\alpha = 4$. This choice only affects the learning rate and does not require tuning. [12]

[21]Indeed, Allen-Zhu and Li [2] indicates that a large rank-$r$ update for the query/value matrices is not crucial. However, a large rank-$r'$ update on the embedding layer is beneficial to address the input distribution shift.

- GPT2(2x), Llama(2x), Mistral(2x) are 6-layer, 24-head, 1536-dim architectures. They are roughly 2x larger than GPT2 small.
- GPT2(5.5x), Llama(5.5x), Mistral(5.5x) are 24-layer, 20-head, 1280-dim architectures. They are roughly 5.5x larger than GPT2 small.[22]

*Remark* B.1. When utilizing the Llama and Mistral architectures, we have also adopted their original tokenizers. It is noteworthy that GPT2's tokenizer converts years (e.g., 19xx) and days into single tokens, whereas the Llama/Mistral tokenizers treat them as four separate tokens. This accounts for certain discrepancies in the partial retrieval accuracies for birth days and birth years.

---

[22]The commercial versions of Llama/Mistral were larger than these and we downsized them for our purpose. For Mistral, we used group-query attention with 4 groups.

(a) the same as Figure 3 but for GPT2 on the bioR datasets

(b) the same as Figure 3 but for Llama on the bioS datasets

(c) the same as Figure 3 but for Llama on the bioR datasets

(d) the same as Figure 3 but for larger GPT2/Llama/Mistral models on 10x to 50x larger bioS datasets

Figure 10: Partial (left) and dual (middle) knowledge retrieval, vs. single knowledge extraction (right).
For descriptions of the datasets (rows), see Appendix A; for architecture and training details, see Appendix B.
**Note:** Unlike real-life QA tasks, our synthetic experiment is trained and fine-tuned on sufficiently clean data for an adequate duration, making it generally unnecessary to increase the model size further; similar results are typically expected.

# C  More Details on Knowledge Classification and Comparison

Recall from Section 4 that we take a model trained on sufficiently augmented BIO data bioS multi5+permute; it is either simply *BIO-pretrained*, denoted as $M$, or already *QA finetuned* on six knowledge extraction QA tasks, denoted as $M'$.[23] We further analyze their performances on knowledge manipulation, particularly on classification or comparison tasks built on certain knowledge attributes.

Consider knowledge comparison as an example. We examine two types of training. One involves direct finetuning of $M$ or $M'$ using manipulation task QAs, such as

> Was Anya Briar Forger born in a month in a year later than Sabrina Eugeo Zuberg? No.

This method is referred to as "train without hint". Once more, we divide the $N$ individuals into two halves $\mathcal{P}_{\text{train}}$ and $\mathcal{P}_{\text{test}}$, apply LoRA fine tuning using QAs for pairs of individuals in $\mathcal{P}_{\text{train}}$, and test its *out-of-distribution* generation accuracy on QAs for pairs of individuals in $\mathcal{P}_{\text{test}}$. We use beam =4 without sampling for model generation (and the results are similar if disabling beam). These results are displayed in the "test acc" column of Figure 4 and 11.

The other training type involves finetuning $M$ or $M'$ using manipulation task QAs *with the addition of hints*, exemplified below:

> Was Anya Briar Forger born in a month in a year later than Sabrina Eugeo Zuberg? October; September. No.

This method enables the model to extract relevant knowledge, then learn to manipulate this knowledge directly. We call this "train with hint", and we again perform LoRA fine tuning using QAs on pairs of individuals in $\mathcal{P}_{\text{train}}$. For each pair of individuals, hints are added with a 50% probability; therefore, during LoRA fine tuning, the model sees knowledge manipulation QAs *both with and without hints*. The model's *out-of-distribution* generation accuracy is then tested on the QAs for individuals in $\mathcal{P}_{\text{test}}$, again with or without hints. These results are displayed in the "test acc (with hint)" and "test acc (w/o hint)" columns of Figure 4 and 11.

Additionally, we document the model's accuracy at correctly generating hints for each individual. This information is presented in the "hint acc" column of Figure 4 and 11.

**Parameters.** The BIO-pretrained model $M$ and QA-finetuned model $M'$ were directly copied from [2]. There were obtained using the same AdamW parameters as described in Appendix B.

Throughout the experiment for both "train without / with hint", we utilize a LoRA finetuning strategy with the rank-16 update on the query/value matrices and rank-128 update on the embedding layer. Additionally, we employ the AdamW optimizer with $\varepsilon = 10^{-6}$. The weight decay is set at 0.01, and the initial learning rate is 0.001. (For the larger Mistral experiment, see below, we use initial learning rate 0.0003 for a better result.) We do not utilize warmup, but we do implement cosine learning rate scheduling, reducing to 10% of the initial learning rate. The batch size is set at 48 with a total of 50,000 training steps. **All the results.** For the GPT2 (12-layer, 12-head, 768-dim) architecture we present our complete results in Figure 11, and a selective set of them in Figure 4 in the main body. Note that not only have we included more classification/ranking/subtraction tasks in Figure 11, but we have also added ranking/subtraction tasks on the birth day attribute, such as "Was [name1] born on a day of the month later than [name2]?" One may note that unlike birth month or major of study, the knowledge of "birth day" can only be retrieved with a less perfect test accuracy of 82.3%. Therefore, one should expect that even with hints added, the knowledge ranking/subtraction accuracy may still be far from perfect. See the last two rows in Figure 4.

We repeat this same experiment for Llama (12-layer, 12-head, 768-dim) in Figure 12 and find the results are almost identical.

---

[23]This QA finetuning is also performed by leveraging LoRA finetuning with rank 8 on the query/value matrices and rank 128 on the embedding layer.

| | | | | BIO pretrained model | | | | QA finetuned model | | | |
|---|---|---|---|---|---|---|---|---|---|---|---|
| | | | baseline | trained w/o hint | trained with hint | | | trained w/o hint | trained with hint | | |
| field | task | #train individuals | | test acc | test acc (with hint) | test acc (w/o hint) | hint acc | test acc | test acc (with hint) | test acc (w/o hint) | hint acc |
| birthmonth | classify %2 | (2.5k) | 50.0 | 60.4 | 77.8 | 65.2 | 64.5 | 61.9 | 80.4 | 65.2 | 69.1 |
| birthmonth | classify %2 | (5k) | 50.0 | 67.3 | 87.3 | 72.7 | 80.3 | 68.0 | 89.5 | 72.8 | 83.9 |
| birthmonth | classify %2 | (10k) | 50.0 | 75.9 | 94.2 | 80.3 | 91.0 | 76.4 | 95.0 | 79.9 | 92.8 |
| birthmonth | classify %2 | (25k) | 50.0 | 86.4 | 98.6 | 91.1 | 97.8 | 87.1 | 98.8 | 90.9 | 98.4 |
| birthmonth | classify %2 | (50k) | 50.0 | 95.3 | 99.5 | 97.5 | 99.2 | 96.3 | 99.7 | 97.5 | 99.5 |
| birthmonth | classify %6 | (2.5k) | 16.7 | 42.1 | 64.4 | 48.6 | 61.8 | 45.9 | 68.0 | 49.8 | 66.0 |
| birthmonth | classify %6 | (5k) | 16.7 | 55.6 | 79.6 | 62.0 | 78.1 | 63.0 | 82.1 | 64.4 | 80.8 |
| birthmonth | classify %6 | (10k) | 16.7 | 76.4 | 90.2 | 75.4 | 89.4 | 79.3 | 92.9 | 78.7 | 92.4 |
| birthmonth | classify %6 | (25k) | 16.7 | 91.9 | 97.5 | 91.5 | 97.2 | 92.8 | 98.5 | 92.1 | 98.4 |
| birthmonth | classify %6 | (50k) | 16.7 | 98.2 | 99.4 | 98.0 | 99.3 | 98.4 | 99.6 | 98.4 | 99.6 |
| birthmonth | classify %12 | (2.5k) | 8.3 | 51.5 | 61.5 | 53.7 | 61.5 | 58.3 | 64.1 | 53.8 | 64.0 |
| birthmonth | classify %12 | (5k) | 8.3 | 74.2 | 79.0 | 70.1 | 79.0 | 80.3 | 82.5 | 75.0 | 82.4 |
| birthmonth | classify %12 | (10k) | 8.3 | 91.6 | 92.0 | 86.8 | 92.0 | 93.5 | 94.7 | 91.2 | 94.7 |
| birthmonth | classify %12 | (25k) | 8.3 | 97.9 | 98.5 | 96.8 | 98.5 | 98.9 | 99.2 | 98.3 | 99.2 |
| birthmonth | classify %12 | (50k) | 8.3 | 99.4 | 99.5 | 99.4 | 99.5 | 99.6 | 99.8 | 99.7 | 99.8 |
| birthmonth | ranking | (2.5k) | 54.2 | 53.7 | 65.4 | 59.6 | 44.2 | 57.3 | 65.5 | 57.6 | 44.9 |
| birthmonth | ranking | (5k) | 54.2 | 59.2 | 75.5 | 63.4 | 63.6 | 62.5 | 75.1 | 63.1 | 62.6 |
| birthmonth | ranking | (10k) | 54.2 | 65.4 | 87.7 | 67.0 | 82.7 | 65.9 | 88.9 | 66.3 | 83.9 |
| birthmonth | ranking | (25k) | 54.2 | 75.6 | 96.7 | 75.8 | 95.4 | 78.3 | 97.4 | 72.5 | 96.3 |
| birthmonth | ranking | (50k) | 54.2 | 85.6 | 99.0 | 86.7 | 98.5 | 88.6 | 98.9 | 82.9 | 98.3 |
| birthmonth | subtraction | (2.5k) | 8.3 | 7.0 | 15.6 | 7.9 | 36.5 | 7.1 | 17.0 | 8.5 | 38.1 |
| birthmonth | subtraction | (5k) | 8.3 | 9.9 | 34.3 | 9.8 | 57.3 | 8.7 | 32.7 | 12.5 | 55.9 |
| birthmonth | subtraction | (10k) | 8.3 | 18.8 | 61.5 | 17.4 | 78.1 | 25.1 | 62.0 | 25.0 | 78.3 |
| birthmonth | subtraction | (25k) | 8.3 | 46.7 | 87.0 | 43.7 | 93.7 | 57.0 | 91.4 | 48.2 | 95.4 |
| birthmonth | subtraction | (50k) | 8.3 | 67.2 | 95.4 | 63.0 | 97.8 | 78.1 | 96.1 | 69.1 | 97.7 |
| major | classify %5 | (10k) | 20.0 | 23.6 | 86.4 | 24.1 | 84.5 | 22.8 | 89.6 | 23.9 | 87.9 |
| major | classify %5 | (25k) | 20.0 | 24.6 | 96.7 | 26.8 | 96.3 | 24.8 | 97.7 | 27.0 | 97.2 |
| major | classify %5 | (50k) | 20.0 | 31.6 | 99.3 | 34.2 | 99.2 | 30.0 | 99.5 | 33.9 | 99.4 |
| major | classify %20 | (10k) | 5.0 | 9.6 | 72.6 | 14.5 | 72.1 | 8.8 | 78.3 | 12.0 | 78.1 |
| major | classify %20 | (25k) | 5.0 | 22.6 | 90.6 | 27.3 | 90.4 | 17.8 | 92.3 | 23.8 | 92.1 |
| major | classify %20 | (50k) | 5.0 | 33.4 | 97.8 | 36.4 | 97.7 | 32.3 | 98.0 | 37.4 | 97.9 |
| major | classify %100 | (10k) | 1.0 | 30.1 | 78.7 | 34.6 | 79.0 | 8.9 | 75.8 | 22.2 | 76.1 |
| major | classify %100 | (25k) | 1.0 | 79.3 | 96.0 | 74.4 | 96.0 | 80.0 | 95.6 | 77.1 | 95.3 |
| major | classify %100 | (50k) | 1.0 | 91.7 | 99.0 | 90.7 | 99.1 | 91.8 | 98.3 | 92.5 | 98.1 |
| major | ranking | (10k) | 50.5 | 52.5 | 88.8 | 54.1 | 86.2 | 52.4 | 90.3 | 54.1 | 88.3 |
| major | ranking | (25k) | 50.5 | 52.2 | 96.4 | 53.7 | 97.3 | 52.6 | 96.9 | 53.6 | 97.5 |
| major | ranking | (50k) | 50.5 | 53.9 | 99.6 | 55.0 | 99.5 | 53.6 | 99.4 | 55.0 | 99.3 |
| major | subtraction | (10k) | 1.0 | 1.1 | 21.6 | 1.1 | 82.5 | 1.0 | 23.2 | 1.1 | 84.3 |
| major | subtraction | (25k) | 1.0 | 1.1 | 89.1 | 1.2 | 96.7 | 1.2 | 84.7 | 1.2 | 97.0 |
| major | subtraction | (50k) | 1.0 | 1.1 | 98.4 | 1.2 | 99.3 | 1.1 | 97.3 | 1.2 | 99.0 |
| birthday | ranking | (50k) | 51.8 | 56.7 | 80.0 | 56.0 | 69.0 | 56.8 | 80.5 | 55.8 | 69.6 |
| birthday | subtraction | (50k) | 3.6 | 4.0 | 45.0 | 4.1 | 68.1 | 4.2 | 45.2 | 4.1 | 69.1 |

Figure 11: An extended version of the GPT2 experiment in Figure 4, to give more examples on knowledge classification and comparison tasks.

We then shoot for a stronger result by using the Mistral (24-layer, 20-head, 1280-dim) in Figure 13 for bioS (50x) dataset (which has $N = 5$ million individuals and even maximum data augmentations, see Remark A.1). Yet, the model is still incapable of learning to compare two majors (among 100 possibilities) when fine-tuned with more than 2.5 million samples — see Figure 13.

| | | | | BIO pretrained model | | | | QA finetuned model | | | |
|---|---|---|---|---|---|---|---|---|---|---|---|
| Llama | | | baseline | trained w/o hint | trained with hint | | | trained w/o hint | trained with hint | | |
| field | task | #train individuals | | test acc | test acc (with hint) | test acc (w/o hint) | hint acc | test acc | test acc (with hint) | test acc (w/o hint) | hint acc |
| birthmonth | classify %2 | (2.5k) | 50.0 | 65.8 | 81.7 | 69.7 | 69.0 | 63.6 | 81.3 | 66.4 | 70.0 |
| birthmonth | classify %2 | (5k) | 50.0 | 77.5 | 89.5 | 81.6 | 81.1 | 78.7 | 89.9 | 80.7 | 83.7 |
| birthmonth | classify %2 | (10k) | 50.0 | 86.3 | 93.8 | 86.9 | 89.0 | 86.1 | 92.2 | 87.8 | 89.4 |
| birthmonth | classify %2 | (25k) | 50.0 | 91.6 | 96.9 | 92.7 | 94.7 | 92.4 | 95.7 | 93.0 | 94.4 |
| birthmonth | classify %2 | (50k) | 50.0 | 95.0 | 98.2 | 95.1 | 96.9 | 95.7 | 99.0 | 96.3 | 98.1 |
| birthmonth | classify %6 | (2.5k) | 16.7 | 60.6 | 68.9 | 60.5 | 64.7 | 56.6 | 73.8 | 64.1 | 71.6 |
| birthmonth | classify %6 | (5k) | 16.7 | 75.2 | 82.1 | 76.1 | 80.3 | 74.3 | 82.7 | 78.3 | 81.5 |
| birthmonth | classify %6 | (10k) | 16.7 | 85.1 | 89.4 | 83.4 | 88.4 | 85.5 | 90.8 | 85.7 | 90.4 |
| birthmonth | classify %6 | (25k) | 16.7 | 92.3 | 94.1 | 90.7 | 92.8 | 92.2 | 92.4 | 91.5 | 92.1 |
| birthmonth | classify %6 | (50k) | 16.7 | 95.0 | 95.6 | 94.6 | 95.0 | 95.8 | 98.0 | 96.4 | 97.9 |
| birthmonth | classify %12 | (2.5k) | 8.3 | 63.8 | 63.1 | 59.9 | 62.9 | 58.7 | 69.7 | 66.1 | 70.2 |
| birthmonth | classify %12 | (5k) | 8.3 | 80.6 | 80.7 | 78.2 | 80.5 | 78.9 | 79.0 | 82.7 | 79.6 |
| birthmonth | classify %12 | (10k) | 8.3 | 89.2 | 88.7 | 86.0 | 88.7 | 89.6 | 80.8 | 89.3 | 86.2 |
| birthmonth | classify %12 | (25k) | 8.3 | 94.6 | 94.6 | 93.1 | 94.6 | 94.9 | 93.9 | 94.7 | 94.5 |
| birthmonth | classify %12 | (50k) | 8.3 | 96.8 | 97.1 | 96.3 | 97.1 | 97.5 | 98.0 | 97.8 | 98.0 |
| birthmonth | ranking | (2.5k) | 54.2 | 52.5 | 65.3 | 57.7 | 45.4 | 52.0 | 65.8 | 55.9 | 46.7 |
| birthmonth | ranking | (5k) | 54.2 | 57.4 | 69.8 | 58.9 | 54.8 | 59.0 | 71.6 | 58.7 | 58.4 |
| birthmonth | ranking | (10k) | 54.2 | 56.8 | 78.8 | 64.9 | 68.3 | 68.1 | 76.9 | 62.2 | 68.1 |
| birthmonth | ranking | (25k) | 54.2 | 61.4 | 90.9 | 75.3 | 86.3 | 75.1 | 90.6 | 76.3 | 87.4 |
| birthmonth | ranking | (50k) | 54.2 | 64.3 | 95.0 | 73.8 | 92.9 | 82.9 | 93.5 | 83.7 | 91.0 |
| birthmonth | subtraction | (2.5k) | 8.3 | 7.3 | 13.3 | 7.7 | 32.6 | 7.9 | 14.3 | 7.3 | 33.8 |
| birthmonth | subtraction | (5k) | 8.3 | 8.0 | 18.4 | 8.0 | 40.6 | 8.8 | 22.1 | 8.0 | 44.8 |
| birthmonth | subtraction | (10k) | 8.3 | 14.6 | 36.1 | 17.8 | 59.1 | 14.0 | 37.8 | 16.6 | 60.7 |
| birthmonth | subtraction | (25k) | 8.3 | 29.7 | 52.1 | 32.5 | 70.3 | 31.7 | 60.5 | 13.7 | 76.3 |
| birthmonth | subtraction | (50k) | 8.3 | 43.7 | 69.0 | 43.4 | 79.6 | 46.0 | 74.6 | 45.3 | 79.9 |
| major | classify %5 | (10k) | 20.0 | 24.2 | 64.1 | 23.8 | 56.3 | 23.0 | 69.1 | 25.4 | 62.6 |
| major | classify %5 | (25k) | 20.0 | 26.9 | 78.0 | 30.4 | 73.2 | 26.4 | 77.6 | 29.7 | 72.9 |
| major | classify %5 | (50k) | 20.0 | 30.6 | 84.1 | 35.8 | 80.9 | 31.1 | 88.7 | 34.6 | 86.7 |
| major | classify %20 | (10k) | 5.0 | 9.3 | 57.3 | 9.7 | 56.8 | 9.9 | 61.2 | 16.1 | 60.1 |
| major | classify %20 | (25k) | 5.0 | 16.7 | 69.3 | 24.2 | 68.2 | 17.4 | 70.0 | 25.7 | 68.7 |
| major | classify %20 | (50k) | 5.0 | 28.2 | 79.0 | 33.1 | 78.3 | 29.9 | 84.2 | 33.2 | 83.5 |
| major | classify %100 | (10k) | 1.0 | 22.3 | 62.0 | 18.9 | 62.4 | 18.5 | 62.1 | 29.7 | 62.2 |
| major | classify %100 | (25k) | 1.0 | 52.7 | 81.7 | 56.3 | 81.9 | 54.9 | 77.9 | 59.1 | 77.9 |
| major | classify %100 | (50k) | 1.0 | 70.5 | 89.2 | 68.9 | 89.3 | 70.5 | 90.7 | 70.5 | 90.8 |
| major | ranking | (10k) | 50.5 | 52.5 | 71.5 | 53.5 | 54.0 | 53.6 | 79.3 | 53.4 | 68.0 |
| major | ranking | (25k) | 50.5 | 54.1 | 88.4 | 53.6 | 84.3 | 54.7 | 89.5 | 54.5 | 85.3 |
| major | ranking | (50k) | 50.5 | 54.6 | 93.0 | 55.4 | 90.9 | 55.6 | 93.5 | 55.7 | 90.8 |
| major | subtraction | (10k) | 1.0 | 0.9 | 12.7 | 1.0 | 62.6 | 0.9 | 14.8 | 1.0 | 64.0 |
| major | subtraction | (25k) | 1.0 | 0.9 | 46.0 | 1.0 | 82.1 | 1.0 | 52.0 | 1.0 | 82.2 |
| major | subtraction | (50k) | 1.0 | 1.0 | 74.3 | 1.0 | 89.0 | 0.9 | 75.6 | 1.1 | 89.8 |
| birthday | ranking | (50k) | 51.8 | 64.0 | 77.0 | 63.8 | 56.1 | 64.5 | 77.3 | 66.3 | 56.0 |
| birthday | subtraction | (50k) | 3.6 | 4.0 | 26.8 | 3.9 | 54.9 | 4.0 | 23.0 | 4.4 | 54.9 |

Figure 12: A repeated experiment of Figure 11 but using Llama architecture of the same size.

| | | | | BIO pretrained model | | | | QA finetuned model | | | |
|---|---|---|---|---|---|---|---|---|---|---|---|
| bioS(50x) | Mistral(5.5x) | | baseline | trained w/o hint | trained with hint | | | trained w/o hint | trained with hint | | |
| field | task | #train individuals | | test acc | test acc (with hint) | test acc (w/o hint) | hint acc | test acc | test acc (with hint) | test acc (w/o hint) | hint acc |
| birthday | ranking | (50k) | 51.8 | 66.7 | 74.4 | 71.5 | 57.5 | 68.9 | 80.1 | 72.7 | 69.6 |
| birthday | ranking | (100k) | 51.8 | 69.2 | 82.0 | 73.6 | 70.6 | 71.2 | 84.4 | 75.2 | 76.4 |
| birthday | ranking | (250k) | 51.8 | 74.6 | 79.9 | 76.6 | 77.0 | 76.9 | 87.7 | 76.9 | 86.3 |
| birthday | ranking | (500k) | 51.8 | 75.0 | 86.5 | 79.7 | 82.1 | 80.1 | 90.8 | 81.3 | 88.0 |
| birthday | ranking | (1m) | 51.8 | 82.3 | 87.9 | 82.6 | 86.9 | 81.8 | 92.3 | 83.7 | 90.4 |
| birthday | ranking | (2.5m) | 51.8 | 87.0 | 94.5 | 86.7 | 93.3 | 88.0 | 95.7 | 87.0 | 93.7 |
| birthday | subtraction | (50k) | 3.6 | 4.4 | 20.1 | 7.0 | 59.4 | 6.3 | 29.9 | 8.4 | 72.2 |
| birthday | subtraction | (100k) | 3.6 | 5.6 | 23.2 | 9.0 | 68.6 | 9.3 | 42.6 | 9.0 | 79.2 |
| birthday | subtraction | (250k) | 3.6 | 12.5 | 44.8 | 14.8 | 76.8 | 14.3 | 51.9 | 17.2 | 85.9 |
| birthday | subtraction | (500k) | 3.6 | 25.0 | 51.6 | 23.5 | 83.9 | 25.3 | 54.6 | 22.8 | 87.5 |
| birthday | subtraction | (1m) | 3.6 | 35.1 | 59.9 | 36.6 | 43.7 | 39.4 | 63.1 | 32.1 | 89.2 |
| birthday | subtraction | (2.5m) | 3.6 | 49.9 | 75.6 | 45.2 | 92.8 | 51.6 | 71.6 | 44.9 | 92.3 |
| major | ranking | (50k) | 50.5 | 51.4 | 65.6 | 57.2 | 46.9 | 52.9 | 73.8 | 57.0 | 57.7 |
| major | ranking | (100k) | 50.5 | 52.2 | 68.4 | 58.9 | 56.1 | 53.6 | 77.3 | 60.1 | 65.3 |
| major | ranking | (250k) | 50.5 | 52.8 | 72.9 | 57.0 | 66.3 | 56.1 | 82.7 | 60.8 | 78.0 |
| major | ranking | (500k) | 50.5 | 56.1 | 81.3 | 60.9 | 74.6 | 59.1 | 84.8 | 60.3 | 77.5 |
| major | ranking | (1m) | 50.5 | 63.8 | 85.1 | 66.7 | 80.8 | 63.0 | 87.8 | 63.1 | 82.9 |
| major | ranking | (2.5m) | 50.5 | 69.7 | 94.4 | 73.2 | 92.5 | 73.0 | 95.7 | 72.0 | 93.7 |
| major | subtraction | (50k) | 1.0 | 1.0 | 9.4 | 1.1 | 46.1 | 1.0 | 14.3 | 1.1 | 61.6 |
| major | subtraction | (100k) | 1.0 | 1.1 | 15.8 | 1.1 | 61.2 | 1.1 | 20.3 | 1.1 | 68.3 |
| major | subtraction | (250k) | 1.0 | 1.1 | 27.4 | 1.1 | 69.4 | 1.0 | 35.4 | 1.2 | 73.1 |
| major | subtraction | (500k) | 1.0 | 1.1 | 52.4 | 1.1 | 81.0 | 1.2 | 49.6 | 1.5 | 77.4 |
| major | subtraction | (1m) | 1.0 | 1.1 | 61.0 | 1.1 | 82.1 | 1.5 | 58.5 | 1.2 | 79.7 |
| major | subtraction | (2.5m) | 1.0 | 1.2 | 78.9 | 1.2 | 90.7 | 3.8 | 82.6 | 1.2 | 91.1 |

Figure 13: A larger experiment than Figure 11, using a 5.5x larger Mistral architecture and 50x training data.

**Observation:** The accuracy of knowledge comparison without CoT remains notably low unless a very large fine-tune dataset is used. For example, the task of subtracting two majors (we have 100 majors, numbered from 0 to 99) cannot be performed better than random guessing even after providing 2.5 million fine-tuning examples. Adding CoTs significantly reduces the required number of samples.

# D  More Details on Knowledge Inverse Search

In Section 5, we examine 10 knowledge inverse search tasks, asking for a person's first or full name given (part or all) of their attributes. We consider the bioS data family with all knowledge augmentation choices as discussed in Appendix A.1.

Similar to knowledge retrieval outlined in Appendix B, given a BIO pretrained model checkpoint, we apply LoRA finetuning on top of it. We do this by utilizing the QA texts of the 10 inverse knowledge search tasks for half of the individuals and test its *out-of-distribution* generation accuracies for answering those QAs on the remaining half. We use the same LoRA and optimization settings as discussed in Appendix B, in particular, rank 8 or 16 for the query/value matrices and rank 128 for the embedding layer, initial learning rate 0.0003, among other parameters. We again use beam =4 without sampling for model generation (and the results are similar if disabling beam).

Furthermore, since we are presenting a negative result, we also consider BIO+QA mixed training. Specifically, we train the model using both the BIO data from all individuals and also the inverse knowledge search QA data from *half* of them. For simplicity, each training sequence of 512 tokens comes either entirely from the BIO entries or entirely from the QA entries (from randomly sampled individuals, concatenated using <EOS> tokens). We introduce a parameter $\mathsf{QA}_r$ to control the frequency of using QA entries. Both $\mathsf{QA}_r = 0.5$ and $\mathsf{QA}_r = 0.8$ are tested, and we present the better result of the two. We evaluate the model's generation accuracy using inverse knowledge search questions from the other half of the individuals.[24]

Our results for the GPT2 (12-layer, 12-head, 768-dim) architecture are in Figure 6. We then repeat this same experiment for Llama (12-layer, 12-head, 768-dim) architecture and Llama tokenizer in Figure 14(a), and the same result holds. We further increased model size and dataset (in the same way as Appendix B) and observed almost identical result in Figure 14(b).

---

[24]As shown in [2], it is deduced that $\mathsf{QA}_r = 0.8$ (specifically, a 2 : 8 ratio between BIO and QA entries in terms of the number of pre-trained tokens) is a good choice for mixed training. However, in the context of inverse knowledge search, the average length of QAs tends to be longer than that of the original knowledge extraction QAs. For this reason, we also explore the alternative option of $\mathsf{QA}_r = 0.5$ to account for this discrepancy.

## Figure 14

| | QA bdate_to_first | QA bdate_to_full | QA b_to_first | QA b_to_full | QA all_to_first | QA all_to_full | QA four_to_first | QA four_to_full | QA three_to_first | QA three_to_full | MIX bdate_to_first | MIX bdate_to_full | MIX b_to_first | MIX b_to_full | MIX all_to_first | MIX all_to_full | MIX four_to_first | MIX four_to_full | MIX three_to_first | MIX three_to_full |
|---|---|---|---|---|---|---|---|---|---|---|---|---|---|---|---|---|---|---|---|---|
| baseline | 0.2 | 0.0 | 0.2 | 0.0 | 0.2 | 0.0 | 0.2 | 0.0 | 0.2 | 0.0 | 0.2 | 0.0 | 0.2 | 0.0 | 0.2 | 0.0 | 0.2 | 0.0 | 0.2 | 0.0 |
| bioS multi5 + reverse1 | 43.7 | 44.0 | 47.8 | 47.2 | 46.2 | 40.9 | 1.0 | 0.0 | 0.8 | 0.0 | 31.2 | 33.1 | 18.4 | 21.6 | 1.0 | 0.6 | 0.7 | 0.0 | 0.8 | 0.0 |
| bioS multi5 + reverse2 | 2.7 | 0.2 | 73.9 | 65.9 | 64.2 | 54.3 | 1.6 | 0.2 | 1.0 | 0.0 | 3.7 | 1.3 | 88.8 | 81.0 | 1.0 | 0.4 | 0.7 | 0.0 | 0.8 | 0.0 |
| bioS multi5 + reverse6 | 1.0 | 0.0 | 1.0 | 0.0 | 62.4 | 57.0 | 53.9 | 37.0 | 42.9 | 28.5 | 0.6 | 0.0 | 0.6 | 0.0 | 95.6 | 91.6 | 95.0 | 92.1 | 90.1 | 88.7 |
| bioS multi5 + permute + reverse6 | 0.6 | 0.0 | 1.2 | 0.0 | 83.6 | 81.5 | 77.6 | 73.1 | 28.0 | 27.2 | 1.0 | 0.0 | 0.8 | 0.2 | 94.1 | 88.5 | 92.2 | 86.8 | 67.0 | 55.0 |
| bioS single | 0.8 | 0.0 | 0.8 | 0.0 | 0.6 | 0.0 | 1.0 | 0.0 | 1.2 | 0.0 | 0.6 | 0.0 | 0.6 | 0.0 | 0.8 | 0.0 | 0.6 | 0.0 | 0.6 | 0.0 |
| bioS single + fullname | 1.0 | 0.0 | 0.8 | 0.0 | 0.6 | 0.0 | 0.8 | 0.0 | 1.2 | 0.0 | 0.8 | 0.2 | 0.8 | 0.0 | 0.8 | 0.0 | 0.7 | 0.0 | 1.0 | 0.0 |
| bioS single + permute1 | 0.8 | 0.0 | 0.6 | 0.0 | 0.6 | 0.0 | 0.8 | 0.0 | 1.0 | 0.0 | 0.6 | 0.0 | 0.8 | 0.0 | 0.8 | 0.0 | 0.7 | 0.0 | 0.8 | 0.0 |
| bioS single + permute2 | 0.8 | 0.0 | 0.8 | 0.0 | 0.8 | 0.0 | 1.2 | 0.0 | 1.0 | 0.0 | 0.6 | 0.0 | 0.8 | 0.0 | 1.0 | 0.0 | 0.7 | 0.0 | 1.0 | 0.0 |
| bioS single + permute5 | 0.8 | 0.0 | 0.6 | 0.0 | 0.6 | 0.0 | 0.6 | 0.0 | 1.2 | 0.0 | 0.6 | 0.0 | 0.6 | 0.0 | 1.0 | 0.0 | 0.7 | 0.0 | 0.8 | 0.0 |
| bioS single + permute1 + fullname | 0.8 | 0.0 | 1.0 | 0.0 | 1.0 | 0.0 | 1.0 | 0.0 | 1.0 | 0.0 | 0.8 | 0.0 | 0.8 | 0.0 | 1.0 | 0.0 | 0.9 | 0.0 | 1.0 | 0.0 |
| bioS single + permute2 + fullname | 0.8 | 0.0 | 0.8 | 0.0 | 0.8 | 0.0 | 0.8 | 0.0 | 1.0 | 0.0 | 0.8 | 0.0 | 0.8 | 0.0 | 0.8 | 0.0 | 0.9 | 0.0 | 1.0 | 0.0 |
| bioS single + permute5 + fullname | 1.0 | 0.0 | 0.8 | 0.0 | 0.6 | 0.0 | 0.6 | 0.0 | 1.0 | 0.0 | 0.6 | 0.2 | 0.6 | 0.0 | 0.8 | 0.0 | 0.7 | 0.0 | 0.8 | 0.0 |
| bioS multi2 | 1.0 | 0.0 | 1.0 | 0.0 | 0.6 | 0.0 | 0.6 | 0.0 | 1.0 | 0.0 | 0.8 | 0.0 | 0.6 | 0.0 | 1.0 | 0.0 | 0.7 | 0.0 | 0.8 | 0.0 |
| bioS multi2 + fullname | 0.8 | 0.0 | 0.6 | 0.0 | 0.6 | 0.0 | 0.6 | 0.0 | 1.0 | 0.0 | 1.0 | 0.2 | 0.6 | 0.0 | 1.0 | 0.0 | 0.6 | 0.0 | 1.0 | 0.0 |
| bioS multi2 + permute | 0.8 | 0.0 | 1.0 | 0.0 | 0.6 | 0.0 | 0.6 | 0.0 | 0.8 | 0.0 | 0.8 | 0.0 | 0.8 | 0.0 | 0.8 | 0.0 | 0.7 | 0.0 | 0.8 | 0.0 |
| bioS multi2 + permute + fullname | 1.3 | 0.0 | 0.8 | 0.0 | 0.6 | 0.0 | 1.0 | 0.0 | 1.0 | 0.0 | 0.8 | 0.0 | 0.8 | 0.0 | 0.8 | 0.0 | 0.6 | 0.0 | 0.8 | 0.0 |
| bioS multi5 | 0.8 | 0.0 | 0.8 | 0.0 | 0.8 | 0.0 | 0.8 | 0.0 | 0.8 | 0.0 | 0.8 | 0.0 | 0.6 | 0.0 | 1.0 | 0.0 | 0.6 | 0.0 | 0.8 | 0.0 |
| bioS multi5 + fullname | 1.0 | 0.0 | 0.6 | 0.0 | 0.8 | 0.0 | 1.2 | 0.0 | 1.0 | 0.0 | 0.8 | 0.2 | 0.8 | 0.0 | 0.8 | 0.0 | 0.7 | 0.0 | 0.8 | 0.0 |
| bioS multi5 + permute | 0.8 | 0.0 | 0.6 | 0.0 | 0.6 | 0.0 | 0.8 | 0.0 | 1.0 | 0.0 | 0.8 | 0.0 | 0.8 | 0.0 | 0.8 | 0.0 | 0.9 | 0.0 | 0.8 | 0.0 |
| bioS multi5 + permute + fullname | 1.0 | 0.0 | 1.0 | 0.0 | 0.8 | 0.0 | 0.6 | 0.0 | 0.8 | 0.0 | 0.8 | 0.0 | 1.0 | 0.2 | 0.8 | 0.0 | 0.9 | 0.0 | 0.8 | 0.0 |

Llama

(a) The same as Figure 6 but using Llama architecture of the same size.

| | QA bdate_to_first | QA bdate_to_full | QA b_to_first | QA b_to_full | QA all_to_first | QA all_to_full | QA four_to_first | QA four_to_full | QA three_to_first | QA three_to_full |
|---|---|---|---|---|---|---|---|---|---|---|
| baseline | 0.2 | 0.0 | 0.2 | 0.0 | 0.2 | 0.0 | 0.2 | 0.0 | 0.2 | 0.0 |
| bioS(10x) multi∞ + permute \| llama(2x) | 0.6 | 0.0 | 0.8 | 0.0 | 1.1 | 0.0 | 0.8 | 0.0 | 0.9 | 0.0 |
| bioS(10x) multi∞ + permute \| gpt2(2x) | 0.9 | 0.0 | 0.6 | 0.0 | 1.2 | 0.0 | 0.6 | 0.0 | 1.0 | 0.0 |
| bioS(10x) multi∞ + permute \| mistral(2x) | 0.6 | 0.0 | 0.6 | 0.0 | 1.4 | 0.0 | 0.8 | 0.0 | 1.1 | 0.0 |
| bioS(20x) multi∞ + permute \| llama(5.5x) | 0.6 | 0.0 | 0.8 | 0.0 | 1.1 | 0.0 | 0.6 | 0.0 | 0.9 | 0.0 |
| bioS(20x) multi∞ + permute \| mistral(5.5x) | 0.6 | 0.0 | 0.8 | 0.0 | 1.4 | 0.0 | 1.0 | 0.0 | 0.9 | 0.0 |
| bioS(20x) multi∞ + permute \| gpt2(5.5x) | 0.9 | 0.0 | 0.6 | 0.0 | 1.0 | 0.0 | 0.6 | 0.0 | 0.8 | 0.0 |
| bioS(50x) multi∞ + permute \| llama(5.5x) | 0.6 | 0.0 | 0.8 | 0.0 | 1.1 | 0.0 | 0.8 | 0.0 | 1.1 | 0.0 |
| bioS(50x) multi∞ + permute \| mistral(5.5x) | 0.6 | 0.0 | 0.8 | 0.0 | 1.4 | 0.0 | 1.2 | 0.0 | 0.9 | 0.0 |
| bioS(50x) multi∞ + permute \| gpt2(5.5x) | 0.9 | 0.0 | 0.6 | 0.0 | 1.2 | 0.0 | 0.6 | 0.0 | 0.8 | 0.0 |

larger models
larger data

(b) Using GPT2/Llama/Mistral of larger sizes and larger data.

Figure 14: We repeat Figure 6 but with more/larger architectures and larger datasets.
For descriptions of the datasets (rows), see Appendix A; for architecture and training details, see Appendix D.

**Note:** Unlike real-life QA tasks, our synthetic experiment is trained and fine-tuned on sufficiently clean data for an adequate duration, making it generally unnecessary to increase the model size further; similar results are typically expected.

# E   More Details on ChatGPT Experiments

All of our experiments on GPT-3.5 / GPT-4 were conducted between June and September of 2023 using the latest models `gpt-3.5-turbo` and `gpt-4` at the moment.

## E.1   Inverse Knowledge Search

In Figure 7 in Section 5, we argued that even massive language models such as GPT-3.5/GPT-4 also perform poorly in inverse knowledge search. We consider four such tasks.

JANE AUSTEN NOVEL TASK.   We select pairs of consecutive sentences in the six novels of Jane Austen, and let GPT-3.5/4 generate the next/previous sentence given the other in the pair. Here, generating the previous sentence can be considered inverse knowledge search, and generating the next sentence can be considered forward knowledge search.

In more detail, we select only those pairs of consecutive sentences when both of them have between 50 and 300 characters (so that we skip short sentences like "What is his name?"). After this filtering, we consider:

- 2873 sentence pairs in *Pride and Prejudice*, out of 5909 sentences;
- 2296 sentence pairs in *Sense and Sensibility*, out of 4897 sentences;
- 2730 sentence pairs in *Persuasion*, out of 3634 sentences;
- 1446 sentence pairs in *Northanger Abbey*, out of 3655 sentences;
- 3234 sentence pairs in *Emma*, out of 8477 sentences;
- 2730 sentence pairs in *Mansfield Park*, out of 6907 sentences.

We then ask GPT3.5/4, "In [bookname ], what's the sentence before/after: [sentence ]?"

WIKIBIO TASK.   We use the wikibio dataset [16], which contains biographies of individuals extracted from Wikipedia. Our goal is to have GPT3.5/4 identify people's names based on their attribute values.

The wikibio dataset consists of 582,659 individuals. We first select only those individuals who have fully specified birth dates, birth places, occupations, and death dates. This results in a total of 33,617 individuals. We then query GPT-3.5 once with the prompt "Answer short: what's the birth day and year of [name ] who is a [occupation ] and was born in [birthplace ]?" and select 4,779 individuals whose birth dates can be corrected answer. This ensures that we only consider individuals that GPT-3.5 has has clearly encountered during its pretraining.

Finally, we test these 4,779 individuals using either GPT-3.5 or GPT-4 with the inverse search question "what's the full name of the celebrity born on [date ] in [city ] who is a [occupation ]?" or the forward search question "what's the birthday and year of [name ] who is a [occupation ] and was born in [city ]?" We assign a score of 1 if the answer is fully correct, and a score of 0.5 if the answer is only partially correct.[25]

CHINESE IDIOM TASK.   We prepared a list of 2,244 four-character Chinese idioms that are commonly used in both oral and written texts. We mask one of the four characters in each idiom and ask GPT3.5/4 to fill in the masked character. In this task, generating the first character given the remaining three characters is considered an inverse knowledge search. Here are a few examples of the idioms that we have used:

---

[25]If only the first or last name is correct, we assign a score of 0.5. If only the birth year is correct, or if both the birth month and day are correct but the year is wrong, we also assign a score of 0.5.

1.实事求是;2.引人注目;3.成千上万;4.当务之急;5.一如既往; ... 2243.秉公守法;2244.等闲置之

We chose to use Chinese because the idioms are of equal length in characters, making it easy to calculate per-character accuracy. An average Chinese individual with a middle school education should be able to achieve an accuracy of over 80% when answering the first character given the other three.

Chinese Poem Task. We prepared a list of 233 Chinese poem sentence pairs that are commonly used in written Chinese. We mask either the first or second sentence and ask GPT-3.5/GPT-4 to complete the other. We provide a few examples of the poem sentence pairs below:

1.两岸猿声啼不住，轻舟已过万重山    2.感时花溅泪，恨别鸟惊心...
... 232.千山鸟飞绝，万径人踪灭    233.东边日出西边雨，道是无晴却有晴

Other tasks. Though we have only presented four tasks related to inverse knowledge search, we have also experimented with a few other tasks not included in the paper. We mention these tasks below for the benefit of interested readers.

- We have tested a wider set of Chinese poems (less frequently used) and Shakespeare's 154 sonnets (which consist of 14 lines of poems each). However, we found that ChatGPT is not very capable at performing even forward search on such tasks. Therefore, it seemed less compelling to test ChatGPT's performance on the corresponding inverse search tasks.

- We have also tested ChatGPT on the Bible, asking it to identify the verse preceding each verse in the same chapter. We found that ChatGPT is capable of performing this task, often with a Chain of Thought (CoT).

  Specifically, remember that the verses in the Bible are properly numbered (for instance, "Gen 15:18" refers to Genesis, chapter 15, verse 18), and the numbers may appear sometimes before and sometimes after the verse. This allows ChatGPT to determine the chapter/verse numbering for a given verse (forward knowledge), perform a "subtract by 1" operation (chain of thought), and then identify the verse using this new number (forward knowledge).

  In other words, we believe the task of asking for the verse preceding each verse in the Bible is actually accomplished by ChatGPT through forward knowledge search + CoT. It is not truly an inverse knowledge search task.

### E.2  Knowledge Classification and Comparison

For knowledge classification and comparison, we once again utilize the pool of 4779 individuals selected from the WikiBio dataset (refer to Section E.1). We then perform the following tasks on GPT-4:

- "Answer me yes or no concisely: for [name ] who was a [occupation ] and was born in [city ] in [year ], was this person born in an even month?"

  We pose this question for every individual in the pool of 4779 people. The baseline accuracy for random guessing in this task is 50%.

- "Answer me yes or no concisely: was [name1 ] who was a [occupation1 ] and was born in [city1 ] born earlier than [name2 ] who was a [occupation2 ] and was born in [city2 ]?"

  We pose this question for 1000 randomly selected pairs of individuals from the pool of 4779 individuals who were either (1) born between 1900-1910, (2) born between 1900-1950, or (3) born in any year. The baseline accuracies for random guessing in these three tasks are: 54.5%, 51.0%, and 50% respectively.

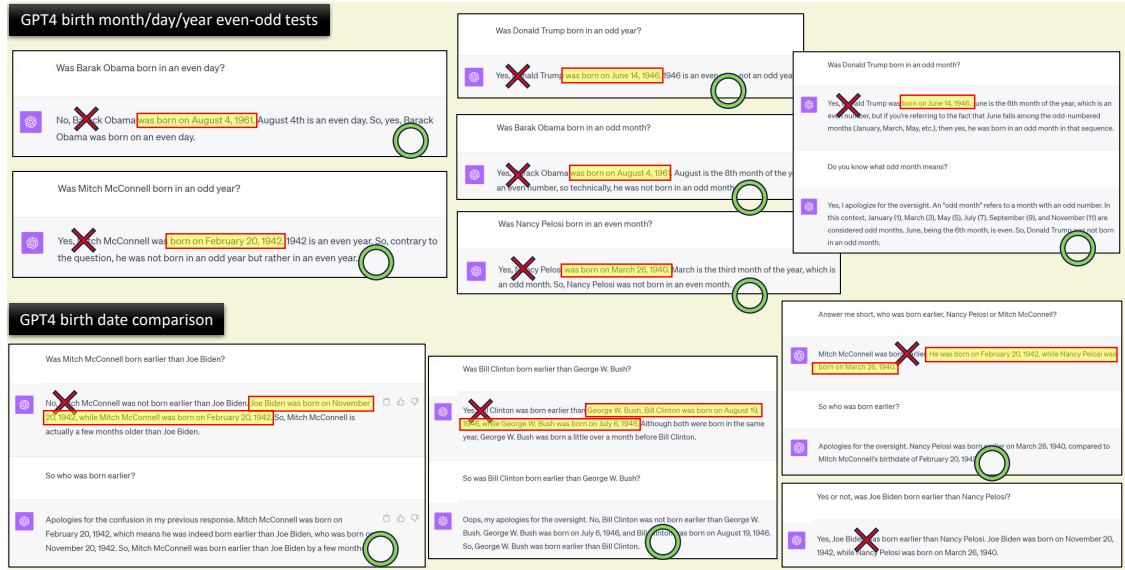

Figure 15: Extension to Figure 2. This figure provides additional examples illustrating GPT-4's difficulty in answering simple manipulation questions based on a person's attributes during inference, despite possessing the necessary knowledge. However, when a Chain of Thoughts (CoT) approach is employed, in which the person's attributes are explicitly stated, GPT-4 is able to correctly answer the manipulation tasks.

Note that in all cases, we prefixed the questions with "answer me yes or no concisely" to compel the model to directly answer with Yes or No without generating a hint first. We present the results in Figure 5.

In addition to the above experiment on WikiBio, we also present some real-life QA examples to illustrate the necessity of the Chain of Thought (CoT). We ask GPT-4 to tell us whether the birth months/days/years of certain politicians are even, as well as to compare the birth dates of some politicians. From the response in Figure 15, it is evident that GPT-4 can easily make mistakes when not using hints (i.e., when answering yes/no without stating the politician's birthdate first), but is capable of correcting such errors once CoT is employed.